\pdfoutput=1
\documentclass{article} % For LaTeX2e

\usepackage[table,xcdraw]{xcolor}
\usepackage{iclr2024_conference,times}

\usepackage{amsmath,amsfonts,bm}

\def\eqref#1{equation~\ref{#1}}
\def\1{\bm{1}}

\DeclareMathAlphabet{\mathsfit}{\encodingdefault}{\sfdefault}{m}{sl}
\SetMathAlphabet{\mathsfit}{bold}{\encodingdefault}{\sfdefault}{bx}{n}

\usepackage[hidelinks]{hyperref}
\usepackage{nicematrix}
\usepackage[capitalize]{cleveref}
\usepackage{times}
\usepackage{latexsym}
\usepackage{mdframed}
\usepackage{microtype}
\usepackage[T1]{fontenc}
\usepackage{tcolorbox}
\usepackage[utf8]{inputenc}

\usepackage{inconsolata}

\usepackage{graphicx}
\usepackage{longtable}
\usepackage{multirow}
\usepackage{wasysym}
\usepackage{enumitem}
\usepackage{hhline}
\usepackage{makecell}
\usepackage{float}
\usepackage{adjustbox}
\usepackage{tikz}
\usepackage{colortbl}
\usepackage{xcolor}
\usepackage{subcaption}

\definecolor{LightCoral}{HTML}{F08080}
\definecolor{LightSeaGreen}{HTML}{20B2AA}
\definecolor{LightSlateGray}{HTML}{778899}
\definecolor{LightGoldenrodYellow}{HTML}{FAFAD2}
\definecolor{Thistle}{HTML}{D8BFD8}
\definecolor{PaleTurquoise}{HTML}{AFEEEE}
\definecolor{Khaki}{HTML}{F0E68C}

\newcommand{\letterR}{
    \tikz[baseline=(char.base)]{
        \node[shape=circle, draw=LightCoral, fill=LightCoral, text=white, inner sep=2pt, font=\sffamily\bfseries] (char) {R};
    }
}
\newcommand{\letterL}{
    \tikz[baseline=(char.base)]{
        \node[shape=circle, draw=LightSeaGreen, fill=LightSeaGreen, text=white, inner sep=2pt, font=\sffamily\bfseries] (char) {L};
    }
}
\newcommand{\letterC}{
    \tikz[baseline=(char.base)]{
        \node[shape=circle, draw=LightSlateGray, fill=LightSlateGray, text=white, inner sep=2pt, font=\sffamily\bfseries] (char) {C};
    }
}
\definecolor{DeepGold}{rgb}{0.72, 0.53, 0.04}
\newcommand{\letterS}{
    \tikz[baseline=(char.base)]{
        \node[shape=circle, draw=DeepGold, fill=DeepGold, text=white, inner sep=2pt, font=\sffamily\bfseries] (char) {S};
    }
}

\newcommand{\letterV}{
    \tikz[baseline=(char.base)]{
        \node[shape=circle, draw=Thistle, fill=Thistle, text=white, inner sep=2pt, font=\sffamily\bfseries] (char) {V};
    }
}
\newcommand{\letterT}{
    \tikz[baseline=(char.base)]{
        \node[shape=circle, draw=PaleTurquoise, fill=PaleTurquoise, text=white, inner sep=2pt, font=\sffamily\bfseries] (char) {T};
    }
}

\definecolor{DeepKhaki}{rgb}{0.6, 0.55, 0.38}
\newcommand{\letterF}{
    \tikz[baseline=(char.base)]{
        \node[shape=circle, draw=DeepKhaki, fill=DeepKhaki, text=white, inner sep=2pt, font=\sffamily\bfseries] (char) {F};
    }
}

\newcommand{\upicon}{
    \raisebox{-0.125em}{\includegraphics[height=1em]{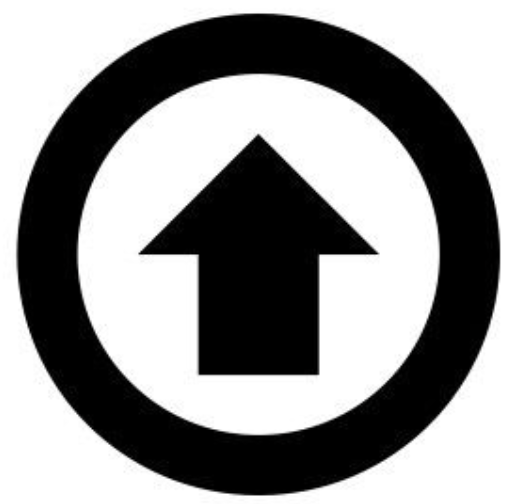}}
}
\newcommand{\equalicon}{
    \raisebox{-0.125em}{\includegraphics[height=1em]{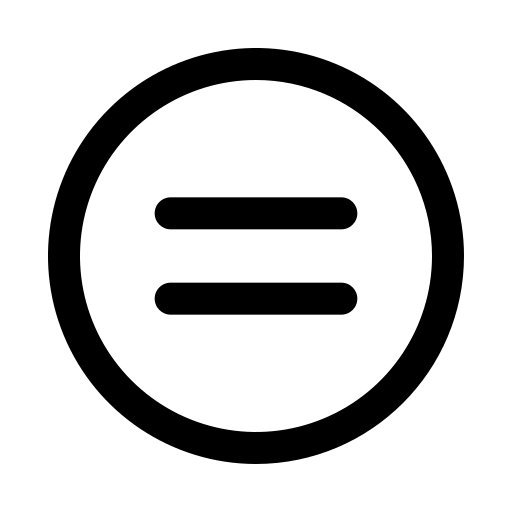}}
}

\title{Evaluating LLMs' Mathematical and Coding Competency through Ontology-guided Interventions}
\newcommand\dataset{\textsc{GSMore}}
\newcommand\datasetcode{\textsc{HumanEval-Core}}

\author{Pengfei Hong$^{1}$, Navonil Majumdar$^1$, Deepanway Ghosal$^{1}$ \\ {\bf Somak Aditya$^2$, Rada Mihalcea$^3$, Soujanya Poria$^1$} \\\\
$^1$ Singapore University of Technology and Design, $^2$ IIT Kharagpur\\
$^3$ University of Michigan\\
}

\usepackage{soul,xcolor}
\usepackage{hyperref}
\usepackage{booktabs}
\usepackage{multirow}
\usepackage{pifont}
\usepackage{listings}
\crefformat{section}{\S#2#1#3} % see manual of cleveref, section 8.2.1
\crefformat{subsection}{\S#2#1#3}
\crefformat{subsubsection}{\S#2#1#3}

\let\realcite\cite
\renewcommand{\cite}[1]{\ifx.#1.\hl{[?]}\else\realcite{#1}\fi}

\iclrfinalcopy
\begin{document}
\maketitle
% \twocolumn[{%
% \renewcommand\twocolumn[1][]{#1}%
% \maketitle
% \begin{center}
%     \vspace{-26pt}
%     \includegraphics[width=\linewidth]{images/llm_robustness.pdf}
%     % \captionof{figure}{\label{fig:teaser}Overview of in-the-wild monologue videos and sentence utterances in the proposed \ds. Each sentence is annotated for $20$ labels including sentiment (and subjectivity), emotions and sentence attributes. ``L'' denotes Likert (intensity) and ``B'' denotes Binary for the type of the labels. The example above is a Portuguese video. \vspace{5mm}}
% \end{center}

\begin{abstract}
Recent advancements in Large Language Models (LLMs) have showcased striking results on existing logical reasoning benchmarks, with some models even surpassing human performance. However, the true depth of their competencies and robustness in reasoning tasks remains an open question. To this end, in this paper, we focus on two popular reasoning tasks: arithmetic reasoning and code generation. 
Particularly, we introduce (i) a general ontology of perturbations for math and coding questions, (ii)  a semi-automatic method to apply these perturbations, and (iii) two datasets, \dataset{} and \datasetcode{}, respectively, of perturbed math and coding problems to probe LLM capabilities in numeric reasoning and coding tasks.
Through comprehensive evaluations of both closed-source and open-source LLMs, we show a significant performance drop across all the models against the perturbed questions, suggesting that the current LLMs lack robust problem solving skills and structured reasoning abilities in many areas, as defined by our ontology. We open-source the datasets and source codes at: \url{https://github.com/declare-lab/LLM-ReasoningTest}.
\end{abstract}

\section{Introduction}

Logical reasoning, that includes mathematics and programming, in a structured and well-defined domain, becomes increasingly complex with the increasing presence of %interspersed and 
diverse situations, events, and contexts represented in natural language. However, the process of solving a mathematical expression should remain consistent and a logical reasoner should be robust to \textit{certain} changes in the context. Consider, problems I) \textit{I have two \underline{apples} and Sam has three more. How many \underline{apples} are there?}, and II) \textit{I see two \underline{wombats}. And my friend sees another three. How many \underline{wombats} are there?}. Variations I and II should yield the same expression ($2+3$), and a similar process to solve resulting in the same answer ($5$). Current state-of-the-art Large Language Models (LLM) have shown impressive performance on a \textit{wide} variety of mathematical problems~\cite{gsm8k} and %reasonable performance on 
coding problems~\cite{chen2021evaluating} expressed in natural language. However, these evaluations barely test the \underline{depth} of LLMs' expertise, and thus we do not currently have fine insights into the LLM capabilities in these domains. %MMLU~\citep{Hendrycks2020MeasuringMM}, MATH~\citep{hendrycks2021measuring}.
For example, in mathematics, GPT-4's performance monotonically decreases from GSM-8k~\cite{gsm8k} ($92\%$; 5-shot CoT) on grade school mathematical problems requiring rigorous arithmetic and logical reasoning to solve; to MMLU-Math ($87.5\%$) \citep{Hendrycks2020MeasuringMM} on a collection of mathematical problems, ranging in difficulty from elementary to advanced levels; and to MATH ($50.36\%$) \citep{hendrycks2021measuring} on challenging competitive mathematics problems. Similar variance in LLM performance can also be observed for coding challenges~\cite{chen2021evaluating}. % \hl{some more can be added here}. 
% However, \textit{can we answer whether GPT-4 can robustly perform grade school-level arithmetic reasoning well irrespective of the complexity of the context?}
Such high-level evaluations show the \textit{breadth} of LLM capabilities, but do not provide any insights into the finer and more fundamental math- and coding-relevant capabilities and shortcomings of LLMs as
(i) many LLMs like GPT-4~\cite{gpt4} are exposed to publicly available math and coding datasets during pre-training, and (ii) many datasets focus on advanced branches of mathematics and problems, skipping the fundamentals. Hence,
% before testing the LLMs' breadth of capabilities by delving into higher-level mathematics and evaluating on competitive coding questions,
we instead focus on depth, encapsulated by one fundamental question: 
\begin{tcolorbox}[colback=green!5, colframe=green!15, sharp corners=all, rounded corners=southwest, title=Research Question, coltitle=black, fonttitle=\bfseries, colbacktitle=green!15]
\textbf{How robust are the capabilities of large language models (LLMs) in reasoning and understanding the problem-solving process?} \end{tcolorbox}

\begin{figure*}[ht]
    %\vspace{-26pt}
    \centering
    \includegraphics[width=\textwidth]{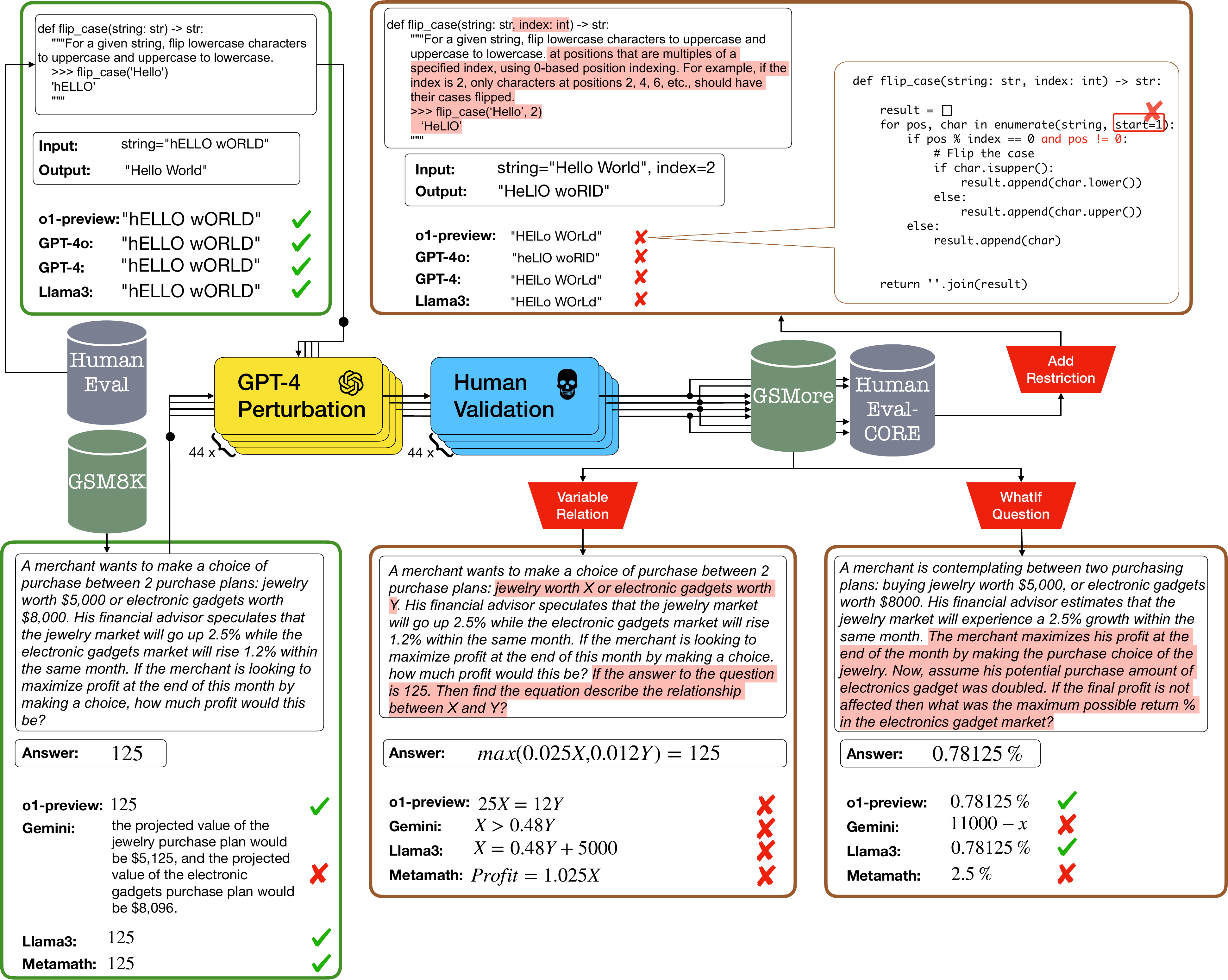}
    \caption{A semi-automated pipeline of creating \dataset{}, from five simple questions from GSM8k. An analogous pipeline is used to create the perturbations of the coding questions from HumanEval, named \datasetcode{}.}
    \label{fig:more_overview}
    % \captionof{figure}{\label{fig:teaser}Overview of in-the-wild monologue videos and sentence utterances in the proposed \ds. Each sentence is annotated for $20$ labels including sentiment (and subjectivity), emotions and sentence attributes. ``L'' denotes Likert (intensity) and ``B'' denotes Binary for the type of the labels. The example above is a Portuguese video. \vspace{5mm}}
\end{figure*}

\begin{tcolorbox}[colback=blue!5, colframe=blue!15, sharp corners=all, rounded corners=southwest, title=Our Approach, coltitle=black, fonttitle=\bfseries, colbacktitle=blue!15]
 Introducing small modifications to the selected seed math problems from GSM8K and HumanEval, where the leading LLMs, such as, o1-preview, have shown excellent performance. Our goal is to examine whether these models can effectively handle slight variations to the seed questions, which they previously solved successfully. An example is presented in \Cref{fig:more_overview}. If the models struggle with these modified problems, it would suggest that there is still room for improvement in LLMs' core reasoning abilities. 
 % Additionally, we explore questions regarding the robustness of their reasoning. 
 \end{tcolorbox}

Evaluating LLMs on these modified questions could provide fine-grained insights into the robustness of the reasoning abilities in the context of math and coding.
% Following previous works on probing language models~\cite{ribeiro2020beyond,wu2023reasoning,Li2024GSMPlusAC, Wang2024BenchmarkSA}, we evaluate the robustness of LLMs' understanding of interesting linguistic and logical structures and derive insights based on them. 
Specifically, our evaluation benchmark is based on our novel ontology of perturbation operations that lists various changes across a diverse set of factors. We apply these perturbations to existing arithmetic and coding problems. These perturbations allow us to assess whether the model truly comprehends underlying concepts and structures or regurgitates. For instance, while a model may correctly answer a math question in a dataset like GSM8k, it may struggle with a simple perturbation to the question, such as replacing numerical values \citep{Srivastava2024FunctionalBF} in math questions with variables, which challenges the model to establish relationships among the variables, revealing its deeper understanding (or lack thereof). By introducing these ontological perturbations, (1) we gain insights into the models' reasoning abilities and (2) uncover strategies for future data augmentation that can then be utilized to enhance LLMs through weakly supervised fine-tuning methodologies.

Our perturbations extend beyond mere numerical alterations to seed questions; they also challenge with the underlying concepts and principles required to solve a problem. For instance, we introduce a category \emph{Critical Thinking}, which includes scenarios where multiple correct answers may be possible, depending on the constraints on various variables, as demonstrated in the example below:

\begin{tcolorbox}
\textbf{Original Question:} \\
A merchant must choose between two purchasing plans: jewelry valued at \$5,000 or electronic gadgets valued at \$8,000. His financial advisor predicts a 2.5\% increase in the jewelry market and a 1.2\% increase in the electronic gadgets market over the next month. If the merchant aims to maximize profit by the end of this month, what would be his profit?\\[0.5em]

\noindent\textbf{Perturbed Question:} \\
A merchant must choose between two purchasing plans: jewelry valued at \$5,000 or electronic gadgets valued at \$8,000. His financial advisor predicts that the jewelry market will rise by \(x\%\) and the electronic gadgets market by 1.2\% over the next month. If the merchant aims to maximize profit by the end of this month, what would his profit be?
\end{tcolorbox}
\noindent Such instances are unique to our datasets making the evaluation more rigorous.

We sample five questions from GSM8K, subjecting them to 44 types of perturbations, followed by extensive manual filtering and validation, resulting in $216$ perturbed questions in \dataset{}. As our ontology is generic, we also apply these perturbations to randomly sampled five code-generation questions from HumanEval producing $219$ questions in \datasetcode{}.
As a whole, this work makes the following contributions:
\begin{enumerate}[itemsep=0pt, leftmargin=*, wide, labelwidth=0pt, labelindent=0pt, parsep=0pt, topsep=0pt]
\item We propose a novel, extensive, and extensible ontology of perturbation operations for math and coding problems.
\item We provide a way to semi-automatically apply such perturbations first through GPT-4, followed by manual filtering. We refer to this dataset as \dataset --- Mathematics-oriented Robustness Evaluation, consisting of $216$ examples starting from just five questions in GSM8K.
\item We show that the ontology and the semi-automatic pipeline can be easily extended to domains like code-generation, by creating the dataset \datasetcode{} --- Coding-oriented Robustness Evaluation, having $219$ samples from $5$ seed questions in HumanEval.
\item We further evaluate the math and code-based reasoning abilities of the latest LLMs, including GPT-4o and o1, using our datasets, revealing their vulnerability to these perturbations.
\item The integration of our ontology with the newly introduced dataset, referred to as \dataset{} and \datasetcode{}, paves the way for a fresh perspective in assessing the math and code reasoning capabilities of LLMs. This evaluation approach transcends the often unidimensional leaderboard performance metrics by delving into LLM performance on finely defined classes of problems.
% A few examples of these challenges include navigating through novel problem constraints, engaging in symbolic reasoning, offering alternative solutions, identifying assumptions, and demonstrating robustness across diverse question formats. This evaluation framework provides a comprehensive understanding of the nuanced areas where LLMs may encounter difficulties, contributing to a more insightful and holistic assessment of their mathematical proficiency.
\end{enumerate}
%Such factors can range from concrete aspects (variable ranges) to abstract ones (what-if scenarios, topics or context of the problem), and cover problem solving aspects, concern instruction following or prompt design and other robustness dimensions. We then ...

\section{Related Work}

\begin{table*}[ht!]
\centering
\resizebox{\textwidth}{!}{
\begin{tabular}{|lccllc|}
\toprule
\textbf{Variant Name} & \textbf{Parent Domain (Dataset)} & \textbf{Type}  & \textbf{Annotation} & \textbf{Dimension} & \textbf{\# Categories} \\ \midrule
SVAMP \citep{Patel2021AreNM} $\star$ & math (ASDiv-A) & Equation-formed list  & Human (Q,A) & \letterV \letterL & 3 \\ 
MetaMathQA \citep{Yu2023MetaMathBY} & math (GSM8K, MATH) & Dynamic CheckList  & GPT-3.5-Turbo & \letterV \letterR & 4 \\ 
GSM-HARD \citep{Gao2022PALPL} & math (GSM8K) & Program-formed CheckList & Codex (Q,A), Human (A) & \letterV & 1 \\ 
GSM-IC \citep{Shi2023LargeLM} $\star$ & math (GSM8K) & Static Checklist  & Human (Q) &  \letterL & 3 \\ 
GSM-PLUS \citep{Li2024GSMPlusAC} $\star$ & math (GSM8K) & Dynamic CheckList  & GPT-4, Human (Q,A) & \letterR \letterL \letterT \letterC &8 \\ 
\dataset{}-\datasetcode{} (Our) $\star$ & \makecell{math (GSM8K), \\ code (HumanEval)} & Dynamic Ontology  & GPT-4, Human (Q,A) & \letterR \letterL \letterC \letterT \letterF \letterS \letterV & 44 \\ 
\bottomrule
\end{tabular}
}
\caption{Overview of variants in reasoning datasets arising from perturbation types. $\star$ refers to datasets specifically designed to evaluate the robustness of model performance. Different letters represent different perturbation types: [R]epresentational Change, [L]ogic Alteration, [C]oncept Analysis, Critical [T]hinking, [F]ormulation Adjustment, [S]caling, [V]alue Replacement.}
\label{tab:related_works}
\end{table*}

A variety of datasets have been developed to assess AI reasoning capabilities across multiple domains. In causal reasoning, significant datasets include those by \citep{Huang2023CLOMOCL, Bondarenko2022CausalQAAB}. For coding, notable contributions have been made by \citep{chen2021codex, Austin2021ProgramSW}. Additionally, mathematical reasoning has been addressed through datasets designed for different educational levels: grade-school \citep{Cobbe2021TrainingVT, gsm8k}, high school \citep{hendrycks2021measuring}, and college level \citep{Sawada2023ARBAR, Zheng2021MiniF2FAC}. Despite the advancements shown by large language models \citep{Ahn2024LargeLM}, recent studies \citep{Mondorf2024BeyondAE} contend that these models more closely resemble stochastic parrots \citep{Bender2021OnTD} than true systematic reasoners, exhibiting significant limitations particularly in scenarios not covered by their training data \citep{Bender2021OnTD, Wan2024AB}.

Therefore, Recent work has focused on the robustness of reasoning under various perturbations that alter reasoning questions. Different domain-specific methods have been proposed for generating test cases for reasoning tasks \citep{Yu2023MetaMathBY, wu2023reasoning}, as summarized in \cref{tab:related_works}. In the field of mathematics, contemporary works have employed techniques such as numerical or symbolic substitutions \citep{Li2024GSMPlusAC, Zhou2023MathAttackAL, Meadows2023ASF, Wang2024BenchmarkSA, Patel2021AreNM}, the insertion of irrelevant distractors \citep{Shi2023LargeLM, Li2023DoYR}, functional equivalence \citep{Srivastava2024FunctionalBF}, and reverse prediction \citep{yu2023metamath, Berglund2023TheRC, Deb2023FillIT} to uncover conceptual errors \citep{Sanyal2022RobustLRAD}, cognitive biases \citep{Dasgupta2022LanguageMS}, or sensitivity to reasoning context \citep{wu2023reasoning}. Similarly, Abstract Syntax Tree guided perturbations are performed on the domain of code \citep{Naik2023SYNCAS, Allamanis2021SelfSupervisedBD}. However, to our knowledge, perturbation methods across the domain of reasoning tasks are lacking. In this work, we consolidate and develop a broader underlying ontology that connects and expands upon previous methods for perturbing reasoning datasets. This new framework is both systematic and hierarchical, and it is readily adaptable to various domains, including mathematics and coding.\\
A distinct line of research focuses on evaluating reasoning through non-conclusion-based assessments, which provide deeper insights into models' reasoning behaviors. For example, ReasonEval \citep{xia2024evaluating} analyzes the \textit{reliability} and \textit{redundancy} of generated reasoning steps, highlighting the qualitative aspects of reasoning. Similarly, \citet{Li2024EvaluatingMR} target at error identification within the reasoning path rather than simply identifying the correct answer. Furthermore, \citet{Zeng2023MRGSM8KAM} explore the robustness of models across varied potential reasoning paths, reinforcing the idea that higher accuracy does not necessarily improve reasoning quality. Our ontology extends these approaches by including perturbations on various concepts related to reasoning path and question understanding, thereby enriching the framework for assessing reasoning capabilities.

\section{The Ontology of Perturbations}
\label{sec:ontology-brief}
\subsection{Need for Ontology of Perturbations} We first identify a set of factors relevant to the solution of a structured reasoning problem expressed in natural language (similar to \citet{kaushik2021learning}); and perturb a seed question under these set of factors semi-automatically in a model-agnostic way (i.e., not necessarily adversarial to a target model). In the NLI context, \citet{kaushik2021learning} utilized human workers to directly perturb a hypothesis, keeping the premise constant; and in a post-hoc way, identifies the categories (or factors) which such revisions pertain to. Previous works~\cite{xu2023wizardlm, Li2024GSMPlusAC, Wang2024BenchmarkSA} discusses ways of perturbation, by identifying a set of factors which is specifically designed to increase the complexity of a seed questions in limited ways. The categories are broad and do not exploit the \textit{logical} nature of the underlying domain (along with the \textit{linguistic} dimensions of the instruction). This is where, we believed, an ontological approach may help, where broader categories can help us generalize, while fine-grained sub-categories exploit the domain-specific characteristics.

Let's take mathematics for example. The solution to a reasoning problem can depend on the number and complexity of operations, variables, functions, and possible existing theorems (external knowledge). Similarly, code generation problems can depend on the data structures, variables, functions, and libraries it needs access to. On top of this well-defined set of factors existing in structured reasoning problems, the list of factors expands as the problem is expressed in natural language. Entities and relations expressed in the text need to be mapped to variables and constants (in both). Physical actions (giving and taking apples) may need to be mapped to mathematical operations (or code). It is clear that the set of \textit{logical} and \textit{linguistic} factors co-exist in these reasoning problems, detailed in \cref{sec:principles_behind_Ontology}.  Therefore we come up with an extensible ontology, capturing the above nuances. We believe it will capture and categorize the factors where LLMs fail over multiple domains. As others have shown, the same process can be enabled to perform data augmentations.

\subsection{The Ontology} Extending SVAMP~\cite{Patel2021AreNM}-like perturbations, we propose a set of high-level categories that are applicable to a broad class of reasoning tasks, expressed in natural language. We primarily identified the following hierarchy (see \cref{tab:new_ontology}):
\\\noindent
\textbf{Level I: Aspect.} There are two aspects to these perturbations: (i) \emph{structural perturbation} and (ii) \emph{representational perturbation}. \emph{Structural perturbation} covers all perturbations that probe the underlying reasoning path (or structure) in different ways, by slightly varying the logic behind the question or probing intermediate steps, seeking explanations. \emph{Representational perturbations} involves modification of the encoding of the question or solution while preserving the underlying logic of the original question.
\\\noindent
% \textbf{Level II: Domain.} The scope of each \emph{aspect} is gradually refined into multiple \emph{domains}. For example, the domain of \emph{logic alteration}, under \emph{structural perturbations}, deals with perturbations that alter the reasoning path in different ways.  
\textbf{Level II: Target.} The subject of change in each \emph{aspect} is gradually refined into multiple \emph{Targets}. For example, the target of \emph{logic}, under \emph{structural perturbations}, deals with perturbations that alter the reasoning path in different controlled ways.  
\\\noindent
\textbf{Level III: Dimension.} This is a further refinement that defines the exact target dimensions (the WHAT) in the reasoning process (question, reasoning, computation, answer expression etc.) to which the perturbations are applied.
\\\noindent
\textbf{Level IV: Category.} This level captures the method (the HOW) through which the higher-level \emph{Dimension} perturbation is achieved. These methods are domain dependent and, thus, their implementations vary from math to coding problems.

\newlength{\aspectlen}
\settowidth{\aspectlen}{\textbf{Representational Perturbation}}
\newlength{\domainlen}
\settowidth{\domainlen}{\textbf{Format Constraint}}
\newlength{\dimlen}
\settowidth{\dimlen}{\textbf{Computation Adjustment}}

\begin{table*}[ht!]
\centering
\small
\resizebox{\textwidth}{!}{
\begin{NiceTabular}{l|l|l|l|l|l}
\toprule
\textbf{Aspect} (Level I) & \textbf{Target} (Level II) & \textbf{Dimension} (Level III) & \textbf{Category} (Level IV) & \textbf{Math} & \textbf{Code} \\ 
\midrule
\multirow{35}{0.6\aspectlen}{
       \textbf{Structural Perturbation} \\
       \textcolor{red}{\textit{Def}: Modification on specific aspects of logic or concepts to alter the reasoning process required to reach the answer}} & \multirow{22}{\domainlen}{\textbf{Logic Alteration}\\
      \textcolor{red}{ \textit{Def}: Modifications to the reasoning framework or logic underpinning a problem.}
       } & \multirow{4}{0.85\dimlen}{\textbf{Granularity Adjustment}\\
      \textcolor{red}{ \textit{Def}: Fine-grained sub-tasks of the original question}
       }  & \ref{gs:remove-constraint} Remove Constraint & Remove Constraint & Remove Constraint \\
& &  & \ref{gs:partial-solution} Partial Solution & Median Inquiry & Helper Function \\
& &  & \ref{gs:solution-plan} Solution Plan & Solution Plan & Solution Plan \\
& &  & \ref{gs:detail-expansion} Detail Expansion & Detail Elaboration & Example Detail \\ 
\cmidrule{3-6} & \\[-1.5ex]
& & \multirow{8}{0.85\dimlen}{\textbf{Reasoning Adjustment}\\
\textcolor{red}{\textit{Def}: Target at logical structure of the original}} % partially change the logical framework in the original problem
& \ref{gs:add-restriction} Add Restriction & Restrict Question & Restrict Requirement \\  
& &  & \ref{gs:subsequent-question} Subsequent Question & Further Question & Further Requirement \\
& &  & \ref{gs:concurrent-question} Concurrent Question & Parallel Question & Parallel Requirement \\
& &  & \ref{gs:change-question} Change Question & Change Query & Change Docstring \\
& &  & \ref{gs:info-recombination} Info Recombination & Info Recombination & Info Recombination \\
& &  & \ref{gs:domain-knowledge} Domain Knowledge & Theoretical Challenge & Code Import \\
& &  & \ref{gs:complex-reality} Complex Reality & Value Probability & Example Boundary \\ 
& &  & \ref{gs:general-solution} General Solution & Code Implementation & Higher Order \\ 
\cmidrule{3-6} & \\[-1.5ex]
& & \multirow{3}{0.85\dimlen}{\textbf{Computation Adjustment}\\
\textcolor{red}{\textit{Def}: Target at values or entities}}  % Inside the logical structure of the original problem
& \ref{gs:computation-demand} Computation Demand & Value Big & Generalize Parameter\\
& &  & \ref{gs:change-value} Change Value & Change Subject & Parameter Content \\ 
& &  & \ref{gs:change-operation} Change Operation & Change Calculation & Variable Type \\  
\cmidrule{3-6} & \\[-1.5ex]
& & \multirow{7}{0.85\dimlen}{\textbf{Formulation Adjustment}\\
\textcolor{red}{\textit{Def}: Reformulate question for solution form to be an abstract expression.}} & \ref{gs:symbolic-response} Symbolic Response & Variable Response & Code Execution \\ 
& &  & \ref{gs:values-relationship} Value Relationship & Variable Relation & Parameter Relationship \\ 
& &  & \ref{gs:variable-group} Variable Group & Variable Scaling & Variable Substitution \\ 
& &  & \ref{gs:backward-reasoning} Backward Reasoning & Variable Adaptation & Reverse Engineering \\ 
& &  & \ref{gs:what-if} Counterfactual & WhatIf Question & WhatIf Code \\ 
& &  & \ref{gs:solve-value} Solve Value & Solve X & Solve Input \\ 
& &  & \ref{gs:identify-range} Identify Range & Variable Range & Variable Range \\ 
\cmidrule{2-6} & \\[-1.5ex]
&\multirow{13}{\domainlen}{\textbf{Concept Analysis}\\
\textcolor{red}{\textit{Def}: Examination and Analysis of the underlying concepts and principles of a problem
}}  & \multirow{4}{0.85\dimlen}{\textbf{Question Understanding}\\
\textcolor{red}{\textit{Def}: Interpretation of the information inside the question}} & \ref{gs:inherent-premise} Inherent Premise & Identify Assumption & Test Case \\ 
& &  & \ref{gs:complete-missing} Complete Missing & Info Sufficiency & Incomplete Answer \\ 
& &  & \ref{gs:question-formulation} Question Formulation & Question Formulation & Question Formulation \\
& &  & \ref{gs:add-misinformation} Add Misinformation & Introduce Distraction & Introduce Bias \\ 
\cmidrule{3-6} & \\[-1.5ex]
& & \multirow{4}{0.85\dimlen}{\textbf{Solution Understanding}\\
\textcolor{red}{\textit{Def}: Assessment of the problem-solving processes}} % within the context of the original problem
& \ref{gs:optimize-solution} Optimize Solution & Info Necessity & Reduce Complexity \\
& &  & \ref{gs:step-functionality} Step Functionality & Step Necessity & Step Necessity \\ 
& &  & \ref{gs:theoretical-basis} Theoretical Basis & Theoretical Basis & Theoretical Basis \\  
& & & \ref{gs:cost-analysis} Cost Analysis & Solution Efficiency & Code Complexity \\
\cmidrule{3-6} & \\[-1.5ex]
&  & \multirow{5}{0.85\dimlen}{\textbf{Critical Thinking}\\
\textcolor{red}{\textit{Def}: Identification of noise, inaccuracies and inconsistencies} }
& \ref{gs:seek-clarification} Seek Clarification & Introduce Ambiguity & Example Requirement \\
& &  & \ref{gs:conditional-analysis} Conditional Analysis & Discuss Separately & Incomplete Requirement \\ 
& &  & \ref{gs:conflicting-information} Conflicting Information & Introduce Contradiction & Wrong Example \\
& &  & \ref{gs:surface-error} Surface Error & Value Uncommon & Runtime Error \\ 
& &  & \ref{gs:hidden-error} Hidden Error & Value Error & Logical Error \\
\midrule
\multirow{12}{0.6\aspectlen}{\textbf{Representational Perturbation}\\
\textcolor{red}{\textit{Def}: Preservation of the underlying logic and conceptual framework, but modification of the encoding or representation
}} & \multirow{7}{\domainlen}{\textbf{Question Format}\\
\textcolor{red}{\textit{Def:} Direct modification on the encoding of the question while keeping the logical structure intact}}  & \multirow{4}{0.85\dimlen}{\textbf{Format Change}\\
\textcolor{red}{\textit{Def}: Rephrasing the question in a different format}}  & \ref{gs:setting-rephrase} Setting Rephrase & Change Setting & Realworld Usecase \\
& &  & \ref{gs:change-sequence} Change Sequence & Change Sequence & Parameter Sequence \\
& &  & \ref{gs:close-format} Close Format & True False & True False \\  
& &  & \ref{gs:data-structuring} Data Restructuring & Value Structuring & Complex Docstring\\ 
\cmidrule{3-6} & \\[-1.5ex]
& & \multirow{3}{0.85\dimlen}{\textbf{Format Comparison}\\
\textcolor{red}{\textit{Def}: Comparing two problem of different forms}} & \multirow{4}{*}{\ref{gs:identical-problem} Identical Problem} & \multirow{4}{*}{Identical Question} & \multirow{4}{*}{Identical Code} \\&&&&\\&&&&\\&&&&\\
\cmidrule{2-6} & \\[-1.5ex]
& \multirow{4}{\domainlen}{\textbf{Answer Format}\\
\textcolor{red}{\texttt{Def}: Indirect modification on the output form}}  
& \multirow{4}{0.85\dimlen}{\textbf{Format Constraint}\\
\textcolor{red}{\textit{Def}: Add constraint on the solution}} & \ref{gs:reasoning-format} Reasoning Format & Binary Coded & No Keyword \\
& &  & \ref{gs:reasoning-style} Reasoning Style & X Language & X Language \\
& &  & \ref{gs:alternative-answer} Alternative Answer & Alternative Answer & Alternative Answer \\
& &  & \ref{gs:new-rule} New Rule & Define Rules & Simple Name\\
\bottomrule
\end{NiceTabular}
}
\caption{Our proposed ontology framework with domain, dimension, mathematical and code realization categories. }
\label{tab:new_ontology}
\end{table*}
\section{Curation of \dataset{} and \datasetcode{}}

\begin{figure*}[t]
  \centering
  \includegraphics[width=\textwidth]{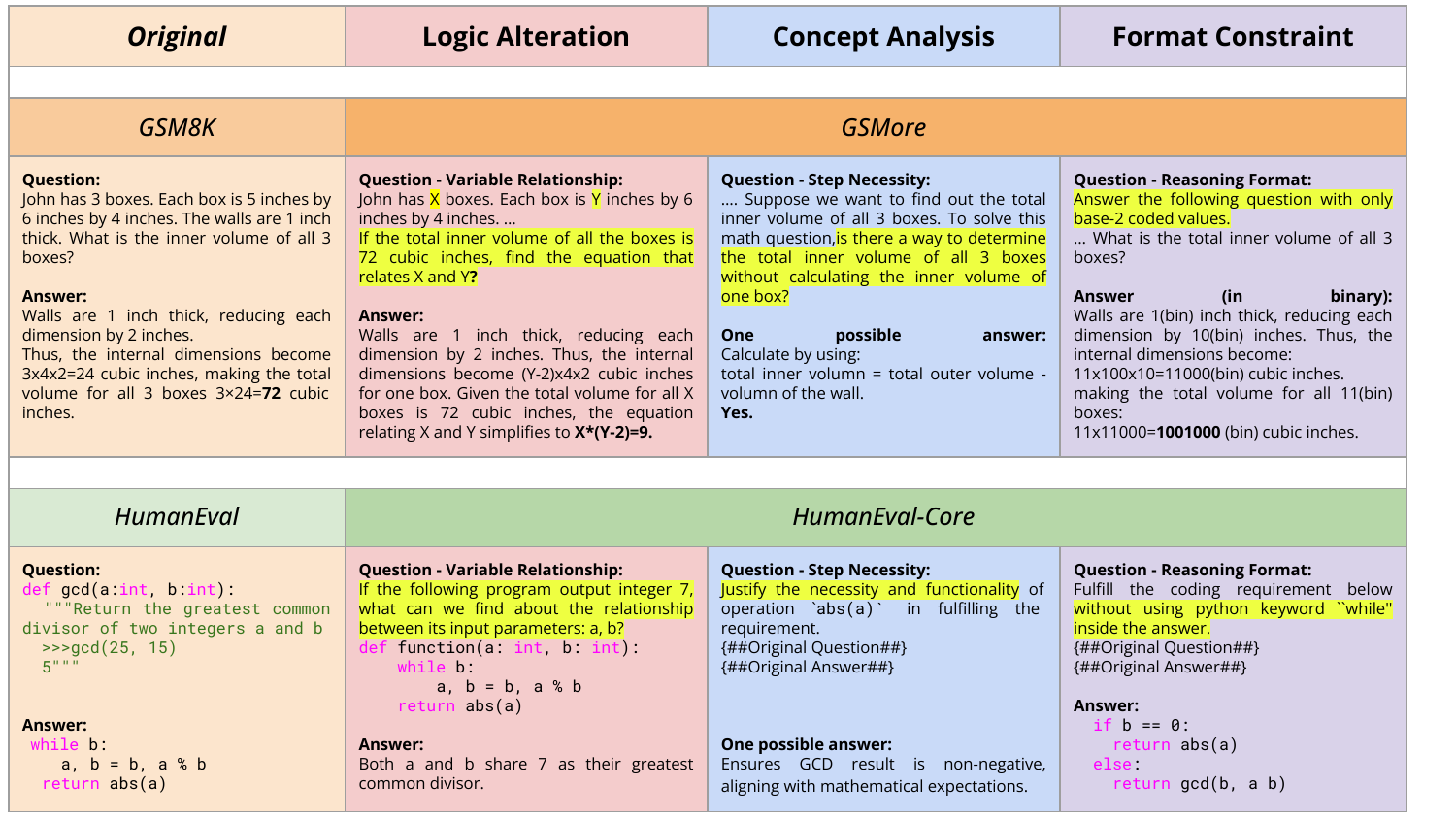}
  \caption{Examples of the original questions and perturbed questions with \emph{Logic, Concept} and \emph{Format} as Targets. The targeted change for each question is highlighted in yellow background.}
  \label{fig:general_criteria}
\end{figure*}

Our objective is to assess the resilience of LLMs to perturbations of math and coding questions along various dimensions. Thus, as seed datasets, we use GSM8K~\citep{Cobbe2021TrainingVT}---a collection of mathematical problems demanding rigorous arithmetic and logical reasoning---and HumanEval~\cite{chen2021evaluating} for coding. Five questions \footnote{Math questions in the GSM8K dataset take between two and eight steps to solve. We randomly chose five questions that take three to seven steps to solve. We cover various topics involving algebraic questions, physical application questions, and decision-based application questions.} from GSM8K are perturbed using our ontological framework (see \Cref{sec:ontology}) to generate \dataset{}. On the other hand, we sampled five coding problems from HumanEval dataset~\citep{chen2021evaluating} that were perturbed using the ontology explained in \Cref{sec:ontology}. These perturbations are aimed at modifying the problems in terms of complexity and representation to assess the robustness of the LLMs to these ontological categories of perturbations.  \cref{fig:general_criteria} shows examples of three perturbed questions and answers from \dataset{} and \datasetcode{}. Examples and definitions of all the remaining perturbations are present in \cref{sec:ontology}. We use a three-staged combination of automatic generation from GPT-4~\citep{gpt4} with human verification and annotation to create \dataset{} and \datasetcode{}: (i) perturbed question generation (\cref{sec:pqg}), (ii) filtering and validation of generated questions (\cref{sec:fav}), and (iii) annotating final answers (\cref{sec:ofa}).

\subsection{Perturbed Question Generation}
\label{sec:pqg}

In the first stage, our objective is to create perturbed questions from the source GSM8K/HumanEval questions for each perturbation type. We write prompt templates for each perturbation type and fill them with a source question to create the input prompt to GPT-4. Each template captures the essence of the respective perturbation type (\Cref{sec:closed-question}, \Cref{sec:open-question}, \Cref{sec:format-change}) to instruct GPT-4 on how to perturb the source question. 
% The template also contains the step-by-step chain of thought solution for the original question. 

For example, the prompt for \textit{Remove Constraint} (\ref{gs:remove-constraint}) for our running example is as follows: 

\begin{tcolorbox}[boxsep=1pt,left=2pt,right=2pt,top=1pt,bottom=1pt,colback=blue!5!white,colframe=gray!75!black]
%\small
Instruction: Rewrite the original mathematical context below based on the \#Rewrite Requirement\#.

Your output should only be \#Rewritten Context\#.

\#Original Context\#: \textcolor{blue}{John has 3 boxes. Each box is 5 inches by 6 inches by 4 inches. The walls are 1 inch thick.}

\#Original Query\#: \textcolor{blue}{ What is the total inner volume of all 3 boxes?}

\#Rewrite Requirement\#: 1. Remove some constraints or information from the original context. 2. Make sure the rewritten question can still be solved, but the answer is simpler.

\#Rewritten Context\#:
\end{tcolorbox}
This prompt to GPT-4 generated: \textit{John has 3 boxes. Each box is 5 inches by 6 inches by 4 inches. What is the total volume of all 3 boxes?}. 
The black text in this prompt marks the static template components to enforce the intended perturbation, while the \textcolor{blue}{blue text} indicates the source question. These templates are used iteratively to generate perturbed questions for GPT-4.

\subsection{Filtering and Validation} 
\label{sec:fav}

% Unfortunately, the GPT-4-generated perturbed questions are not always meaningful and suitable for robustness testing. This is expected as many perturbation types are quite complex and open-ended for GPT-4 to make a mistake in the generation. Furthermore, as also shown in \citet{Li2024GSMPlusAC}, GPT-4 may (i) fail to incorporate perturbations, e.g., for \emph{Data Restructuring}, it fails to identify all the values in the question (ii) include additional changes beyond the specified perturbations.
% % incapable of meticulously following the given instructions to always generate clear and logically-coherent perturbed questions. These factors often lead to unsatisfactory perturbations that do not capture the essence of the perturbation types. 
% In particular, we would like to maintain the following properties: Human Understandability, Logical Coherence, Instruction Adherence as detailed in Appendix \ref{sec:properties}

% Hence, we implement a semi-automatic filtering process to ensure the above qualities and relevance of the generations. We first perform an automated filtering process where GPT-4 itself determines whether a generated question meets all the above three criteria. We subsequently discard all questions that fail to fulfill any of the three criteria. For these instances, we regenerate a new perturbed question from GPT-4 and check if it clears the filtering process. If it fails again then we ask a human annotator to write the perturbed question. 
Unfortunately, GPT-4-generated perturbed questions sometimes lack meaning and suitability for robustness testing due to complex and open-ended perturbation types, leading to errors in generation. As noted in \citet{Li2024GSMPlusAC}, GPT-4 may i) fail to incorporate perturbations, such as missing values in \emph{Data Restructuring}, ii) introduce unintended changes. We aim to maintain Human Understandability, Logical Coherence, and Instruction Adherence, as detailed in Appendix \ref{sec:properties}.

% To ensure these qualities and relevance, we use a semi-automatic filtering process. Initially, GPT-4 performs an automated check against the three criteria, discarding any questions that do not meet them. Failed questions are regenerated and re-evaluated, with persistent failures handled by a human annotator.
To ensure these qualities and relevance, we use a semi-automatic filtering process. Initially, GPT-4 performs an automated check against the three criteria, discarding any questions that do not meet them. Questions that repeatedly fail are then reviewed and handled by a human annotator. 

\paragraph{Human Verification.} 
% We still expect the perturbed questions obtained after filtering to have some limitations, as automatic verification is not a perfect process. Hence, we do a final round of human verification to rephrase/rewrite the perturbed questions to make them clean and correct. We found that 36\% of the filtered questions were nearly correct and only required some minor rewording. However, 31\% of the filtered questions had either significant inaccuracies or failed the filtering process. We ask human annotators to write the correct perturbed question for these instances. The other 33\% of the filtered questions were correct and did not require any further revision. 
% % We observed the inaccuracies in terms of i) insufficient question information, ii) logical incorrectness, and iii) inappropriate perturbation type.

% We thus ensure that the final set of questions in \dataset{} are of high quality, understandable, logically coherent, and in line with the intended perturbation method.

% The human verification is conducted by five annotators, who all have strong mathematical foundations as PhD students in the field of computer science. Each rephrase/rewrite was performed by one annotator, whose judgment was then verified by the other two annotators.

Despite automatic verification, perturbed questions still have limitations, so we conduct a final human verification to refine them. Our findings show that 36\% of the filtered questions needed minor rewording, 31\% contained significant inaccuracies or failed the filtering, and 33\% were correct as is. Thus, the final questions in \dataset{} and \datasetcode{} are high-quality, understandable, logically coherent, and aligned with the intended perturbation method. Human verification is performed by five PhD computer science students, with each question revised by two annotators and verified by two others.

\subsection{Obtaining Final Answers of the Perturbed Questions}
\label{sec:ofa}

Finally, we also annotate the gold answer for the perturbed questions. We engaged the same five annotators for this process. Each gold answer was initially annotated by one annotator. Subsequently, the annotated responses underwent verification by the other two annotators.

\subsection{Statistics of \dataset{} and \datasetcode{}}

We sampled five questions from GSM8K and HumanEval and perturbed them using GPT-4 in 44 distinct perturbation categories. Following a rigorous process of filtering and validation, we retained a total of 216 and 219 perturbed questions in \dataset{} and \datasetcode{}, respectively.  We specify the detailed statistics in Appendix \ref{sec:dataset_details} and the details of the five selected questions from each dataset in Appendix \ref{app:original_gsm8k} and Appendix \ref{app:original_humaneval} respectively.

\section{Experiments}

\subsection{Evaluation Protocol}

Owing to the loosely controlled format of the LLM responses to the majority of the questions, calculating accuracy through direct string matching with the annotated answer may not always be reliable. Additionally, in the context of \emph{concept analysis}, curating an exhaustive list of correct answers could be intractable. For instance, the category \emph{optimize solution} (\ref{gs:optimize-solution}) asks to further optimize the provided solution. There could be numerous distinct valid ways to optimize the given solution. To address these challenges, we manually evaluated the generated outputs. For \datasetcode{}, we designed test cases where possible. The evaluation is fully automated for those instances. For the remaining instances without test cases, we manually verified the answers returned by the LLMs.

%To empirically justify this, we prompted GPT-4 for automated answer evaluation, yielding an agreement of 88.76\% with human annotation on the answers of GPT-4 to \dataset{} questions.  

\paragraph{Automated Evaluation.}
With the recent advancements in the reasoning abilities of large language models (LLMs), we also utilize GPT-4o as an evaluator to assess the correctness of answers against the manually annotated ground truth. GPT-4o proved to be an effective evaluator, showing a high level of agreement with human judgments across various models. We report the automated evaluation results and their agreement with manual evaluation in \Cref{table:main_result2}.

\subsection{Experimental Setup}

We evaluated five prominent closed- and open-sourced LLMs on our benchmark. The closed-sourced LLMs are o1-preview, GPT-4o, GPT-4, GPT-3.5, and Gemini-1.5-pro. The remaining open-sourced LLMs include one general-purpose LLM and one LLM finetuned on task-specific datasets. The general-purpose LLM is Llama3-8B-Instruct and task-specific LLMs are MetaMath-70B-V1.0 and CodeLlama-70B-Instruct for coding and math, respectively. MetaMath-70B-V1.0 is finetuned on a mixture of datasets from Metamath \citep{yu2023metamath} and Mistral \citep{Jiang2023Mistral7} and CodeLlama-70B-Instruct is finetuned on publicly available coding and coding-related instructions~\cite{Rozire2023CodeLO}. Model Details are specified in Appendix~\ref{sec:model_details}.  We listed the prompts used for these models in Appendix~\ref{app:eval-dets}. Each question is evaluated with \texttt{pass@1} metric under zero-shot setting. More details are in Appendix~\ref{app:eval-dets} on the evaluation settings.

\subsection{Experimental Results and Analyses}

\begin{table*}[ht!]
\centering
\resizebox{\textwidth}{!}{%
\begin{NiceTabular}{l|c|cc|ccc>{\columncolor{Thistle}}cccc>{\columncolor{Thistle}}ccc>{\columncolor{Thistle}}ccc|c}
\toprule
&  & \multicolumn{2}{c|}{Original} & \multicolumn{9}{c}{Structural} & \multicolumn{4}{c|}{Representational} &   \\
\cmidrule(lr){5-13} \cmidrule(lr){14-17} 
&  & & & \multicolumn{5}{c}{Logic Alteration} & \multicolumn{4}{c}{Concept Analysis} & \multicolumn{3}{c}{Question Format} & A. Format & \it Weighted  \\
\cmidrule(lr){3-4}\cmidrule(lr){5-9} \cmidrule(lr){10-13} \cmidrule(lr){14-16} \cmidrule(lr){17-17}
 & \multirow{2}{*}{Models}& Seed & All & Gran. & Reason & Compute. & Formul & \it Avg. & Quest. & Sol. & Crit. & \it Avg. & Form. & Form. & \it Avg. & Form. & \it Avg. \\
& & Samp. & Samp. & Adjust. & Adjust. & Adjust. & Adjust. & \it Perf. & Under. & Under. & Think. & \it Perf. & Change. & Comp. & \it Perf. & Constraint \\
\midrule
\multirow{6}{*}{\rotatebox[origin=c]{90}{MATH}}
& o1-preview & 100.0 & 95.2 & 100.0 & 92.5 & 92.3 & 77.1 & 88.9 & 95.0 & 95.0 & 56.0 & 80.0 & 95.0 & 80.0 & 92.0 & 90.0 & 86.7 \\
& GPT-4o & 100.0 & 95.1 & 100.0 & 92.5 & 81.8 & 71.4 & 85.8 & 95.0 & 90.0 & 48.0 & 75.4 & 100.0 & 80.0 & 96.0 & 85.0 & 83.8 \\
& GPT-4 & 100.0 & 94.6 & 100.0 & 80.0 & 90.9 & 60.0 & 78.3 & 85.0 & 65.0 & 48.0 & 64.6 & 90.0 & 60.0 & 84.0 & 65.0 & 74.2 \\
& GPT-3.5 & 80.0 & 78.9 & 75.0 & 27.5 & 54.6 & 25.7 & 38.7 & 55.0 & 45.0 & 12.0 & 35.4 & 35.0 & 40.0 & 36.0 & 5.0 & 35.8 \\
& Gemini & 80.0 & 91.7 & 90.0 & 50.0 & 81.8 & 37.1 & 56.6 & 60.0 & 20.0 & 16.0 & 30.8 & 55.0 & 20.0 & 48.0 & 30.0 & 46.2 \\
& Llama3 & 60.0 & 80.7 & 50.0 & 12.5 & 18.2 & 5.7 & 17.9 & 35.0 & 60.0 & 4.0 & 30.8 & 5.0 & 60.0 & 16.0 & 5.0 & 26.2 \\
& Metamath & 80.0 & 82.3 & 70.0 & 15.0 & 27.3 & 11.4 & 25.5 & 30.0 & 25.0 & 4.0 & 18.5 & 35.0 & 80.0 & 44.0 & 20.0 & 21.3 \\
\cmidrule{2-18}

% & Average & 80 &  \hl{--}& 77 & 37 & 54.55 & 27.90 & 43.39 & 53 & 43 & 16.8 & 36.00 & 44 & 52 & 45.60 & 25 & 40.72 \\
& Average & 85.7 & 88.4 & 83.6 & 52.9 & 63.8 & 41.2 & 56.0 & 65.0 & 57.1 & 26.9 & 47.9 & 59.3 & 60.0 & 59.4 & 42.9 & 53.5 \\
\noalign{\vskip 1.5mm}
\Xhline{1.25pt}
\noalign{\vskip 1.5mm}
\multirow{6}{*}{\rotatebox[origin=c]{90}{CODE}} 
& o1-preview & 100.0 & 92.4 & 95.0 & 62.5 & 46.7 & 80.0 & 71.8 & 95.0 & 85.0 & 64.0 & 80.0 & 75.0 & 80.0 & 76.0 & 95.0 & 76.8 \\
& GPT-4o & 80.0 & 90.2 & 90.0 & 57.5 & 53.3 & 74.3 & 68.1 & 80.0 & 95.0 & 56.0 & 75.4 & 65.0 & 60.0 & 64.0 & 70.0 & 70.0 \\
& GPT-4 & 80.0 & 76.5 & 90.0 & 37.5 & 46.7 & 50.0 & 52.3 & 65.0 & 80.0 & 44.0 & 61.5 & 65.0 & 40.0 & 60.0 & 55.0 & 56.7 \\
& GPT-3.5 & 80.0 & 64.9 & 73.7 & 35.0 & 40.0 & 29.4 & 40.7 & 60.0 & 75.0 & 40.0 & 56.9 & 50.0 & 40.0 & 48.0 & 45.0 & 47.1 \\
& Gemini & 80.0 & 71.9 & 80.0 & 32.5 & 53.3 & 23.5 & 41.3 & 65.0 & 75.0 & 44.0 & 60.0 & 45.0 & 40.0 & 44.0 & 35.0 & 47.3 \\
& Llama3 & 60.0 & 62.2 & 45.0 & 12.5 & 33.3 & 11.8 & 21.1 & 50.0 & 50.0 & 8.0 & 33.9 & 25.0 & 40.0 & 28.0 & 20.0 & 36.6 \\
& CodeLlama & 60.0 & 67.8 & 80.0 & 40.0 & 40.0 & 11.8 & 38.5 & 35.0 & 35.0 & 28.0 & 32.3 & 40.0 & 0.0 & 32.0 & 40.0 & 26.3 \\
\cmidrule{2-18}

% & Average & 72 & 72.85 & 73.74 & 31.5 & 42.67 & 25.29 & 38.79 & 55 & 63 & 32.8 & 48.92 & 45 & 32 & 42.40 & 39.00 & 42.81 \\
& Average & 77.1 & 75.1 & 79.1 & 39.6 & 44.7 & 40.1 & 47.7 & 64.3 & 70.7 & 40.6 & 57.1 & 52.1 & 42.9 & 50.3 & 51.4 & 51.5 \\
\bottomrule
\end{NiceTabular}
}
\caption{Model performance on math and coding across various \emph{Dimensions} (Level III of ontology). All the average reported is weighted average. }
\label{table:main_result}
\end{table*}

\begin{table*}[ht!]
\centering
\resizebox{\textwidth}{!}{%
\begin{NiceTabular}{l|c|cc|ccc>{\columncolor{Thistle}}cccc>{\columncolor{Thistle}}ccc>{\columncolor{Thistle}}ccc|c|c}
\toprule
&  & \multicolumn{2}{c|}{Original} & \multicolumn{9}{c}{Structural} & \multicolumn{4}{c|}{Representational} &  &  \\
\cmidrule(lr){5-13} \cmidrule(lr){14-17} 
&  & & & \multicolumn{5}{c}{Logic Alteration} & \multicolumn{4}{c}{Concept Analysis} & \multicolumn{3}{c}{Question Format} & A. Format & \it Weighted  & Auto-eval  \\
\cmidrule(lr){3-4}\cmidrule(lr){5-9} \cmidrule(lr){10-13} \cmidrule(lr){14-16} \cmidrule(lr){17-17}
 & \multirow{2}{*}{Models}& Seed & All & Gran. & Reason & Compute. & Formul & \it Avg. & Quest. & Sol. & Crit. & \it Avg. & Form. & Form. & \it Avg. & Form. & \it Avg. & Agreement \\
& & Samp. & Samp. & Adjust. & Adjust. & Adjust. & Adjust. & \it Perf. & Under. & Under. & Think. & \it Perf. & Change. & Comp. & \it Perf. & Constraint &  &  \\
\midrule
\multirow{6}{*}{\rotatebox[origin=c]{90}{MATH}}
& o1-preview & 100.0 & 95.2 & 95.0 & 90.0 & 91.7 & 62.9 & 82.2 & 75.0 & 50.0 & 20.0 & 46.2 & 85.0 & 80.0 & 84.0 & 85.0 & 71.9 & 84.3 \\
& GPT-4o & 100.0 & 95.1 & 100.0 & 92.5 & 81.8 & 71.4 & 85.8 & 95.0 & 90.0 & 48.0 & 75.4 & 100.0 & 80.0 & 96.0 & 85.0 & 71.3 & 85.6 \\
& GPT-4 & 100.0 & 94.6 & 90.0 & 67.5 & 90.9  & 51.4 & 68.9 & 45.0 & 15.0 & 24.0 & 27.7 & 85.0 & 40.0 & 76.0 & 80.0 & 58.3 & 81.0 \\
& GPT-3.5 & 80.0 & 78.9 & 75.0 & 25.0 & 54.5 & 14.3 & 34.0 & 50.0 & 30.0 & 8.0 & 27.7 & 40.0 & 60.0 & 44.0 & 25.0 & 32.4 & 85.2 \\
& Gemini & 80.0 & 91.7 & 80.0 & 60.0 & 81.8 & 31.4 & 56.6 & 45.0 & 10.0 & 12.0 & 21.5 & 45.0 & 40.0 & 44.0 & 70.0 & 45.8 & 86.6 \\
& Llama3 & 60.0 & 80.7 & 65.0 & 37.5 & 45.5 & 14.3 & 35.8 & 15.0 & 15.0 & 8.0 & 12.3 & 30.0 & 40.0 & 32.0 & 30.0 & 27.8 & 88.9 \\
& Metamath & 80.0 & 82.3 & 75.0 & 22.5 & 45.5 & 14.3 & 32.1 & 30.0 & 10.0 & 8.0 & 15.4 & 40.0 & 100.0 & 52.0 & 45.0 & 30.6 & 89.8 \\
\cmidrule{2-19}

& Average & 85.7 & 88.4 & 82.9 & 56.4 & 70.2 & 37.1 & 56.5 & 50.7 & 31.4 & 18.3 & 32.3 & 60.7 & 62.9 & 61.1 & 60.0 & 48.3 & 85.9 \\

\noalign{\vskip 1.5mm}
\Xhline{1.25pt}
\noalign{\vskip 1.5mm}
\multirow{6}{*}{\rotatebox[origin=c]{90}{CODE}} 
& o1-preview & 100.0 & 92.4 & 90.0 & 55.0 & 46.7 & 71.4 & 58.5 & 70.0 & 60.0 & 48.0 & 65.5 & 70.0 & 80.0 & 72.0 & 85.0 & 65.9 & 86.4 \\
& GPT-4o & 80.0 & 90.2 & 90.0 & 47.5 & 46.7 & 65.7 & 60.9 & 65.0 & 60.0 & 32.0 & 50.7 & 60.0 & 40.0 & 56.0 & 70.0 & 58.2 & 85.5 \\
& GPT-4 & 80.0 & 76.5 & 100 & 37.5 & 46.7 & 50.0 & 52.3 & 75.0 & 65.0 & 48.0 & 61.5 & 65.0 & 40.0 & 60.0 & 55.0 & 74.5 & 71.7 \\
& GPT-3.5 & 80.0 & 64.9 & 73.7 & 35.0 & 40.0 & 29.4 & 40.7 & 55.0 & 70.0 & 40.0 & 56.9 & 50.0 & 40.0 & 48.0 & 45.0 & 46.3 & 87.2 \\
& Gemini & 80.0 & 71.9 & 80.0 & 32.5 & 53.3 & 23.5 & 41.3 & 65.0 & 55.0 & 24.0 & 46.2 & 45.0 & 40.0 & 44.0 & 35.0 & 43.2 & 85.4 \\
& Llama3 & 60.0 & 62.2 & 45.0 & 12.5 & 33.3 & 11.8 & 21.1 & 50.0 & 40.0 & 0.0 & 46.1 & 25.0 & 40.0 & 28.0 & 25.0 & 36.6 &  91.8 \\
& CodeLlama & 60.0 & 67.8 & 80.0 & 40.0 & 40.0 & 11.8 & 38.5 & 25.0 & 15.0 & 16.0 & 18.0 & 40.0 & 0.0 & 32.0 & 40.0 & 26.4 & 79.0 \\
\cmidrule{2-19}

& Average & 77.1 & 75.1 & 79.8 & 37.1 & 43.8 & 37.7 & 44.8 & 57.9 & 52.1 & 29.7 & 49.3 & 50.7 & 40.0 & 48.6 & 50.7 & 50.2 & 83.9 \\
\bottomrule
\end{NiceTabular}
}
\caption{Model performance on math using automated evaluation across various \emph{Dimensions} (Level III of ontology). All the average reported is weighted average. }
\label{table:main_result2}
\end{table*}

\begin{table}[ht]
\centering
\scriptsize
\begin{NiceTabular}{ll|cccc|c}
\toprule
 & \multirow{2}{*}{Models} & G. & R. & C. & F. & \multirow{2}{*}{\textit{Avg.}} \\
 & & Adj. &  Adj. & Adj. &  Adj. & \\
\midrule
\multirow{3}{*}{\rotatebox[origin=c]{90}{\scriptsize MATH }} & GPT-4 & 100 & 77.5 & 90.91 & 71.43 & 81.13 \\ 
& GPT-3.5 & 90 & 50 & 90.91 & 40 & 58.49 \\
& Gemini & 95 & 57.50 & 63.64 & 45.71 & 61.32 \\
\noalign{\vskip 1.5mm}
\Xhline{1.25pt}
\noalign{\vskip 1.5mm}
\multirow{3}{*}{\rotatebox[origin=c]{90}{\scriptsize CODE }} & GPT-4 & 100 & 50 & 46.67 & 55.88 & 59.43 \\
& GPT-3.5 & 82.35 & 42.50 & 40 & 26.47 & 43.40 \\
& Gemini & 70.59 & 25 & 53.33 & 26.47 & 36.79 \\
\bottomrule
\end{NiceTabular}
\caption{The impact of incorporating the original question and answer into the prompt on the performance of \emph{logic} Target within the \dataset{} and \datasetcode{}. The reported average is weighted average.}
\label{table:logic_alteration}
\end{table}

\begin{table}[ht]
\centering
\scriptsize
\begin{NiceTabular}{ll|cccc|c}
\toprule
 & \multirow{2}{*}{Models} & G. & R. & C. & F. & \multirow{2}{*}{\textit{Avg.}} \\
 & & Adj. &  Adj. & Adj. &  Adj. & \\
\midrule
\multirow{3}{*}{\rotatebox[origin=c]{90}{\scriptsize Self-C }} & GPT-4 & 95 & 87.5 & 90.91 & 65.71 & 82.08 \\ 
& GPT-3.5 & 60 & 45 & 45.45 & 25.71 & 41.51 \\
& Gemini & 75 & 45 & 81.82 & 40 & 52.83 \\
\noalign{\vskip 1.5mm}
\Xhline{1.25pt}
\noalign{\vskip 1.5mm}
\multirow{3}{*}{\rotatebox[origin=c]{90}{\scriptsize POT }} & GPT-4 & 95 & 90 & 81.82 & 68.57 & 83.02 \\
& GPT-3.5 & 75 & 57.5 & 54.55 & 25.71 & 50 \\
& Gemini & 90 & 60 & 63.64 & 45.71 & 61.32 \\
\bottomrule
\end{NiceTabular}
\caption{The impact of using prompting techniques on the performance of \emph{Logic} Target within the \dataset{} and \datasetcode{}. Self-C stands for Self-Consistency prompting \citep{Wang2022SelfConsistencyIC} and POT stands for Program of Thought \cite{Chen2022ProgramOT}.}
\label{table:logic_alteration_prompting}
\end{table}

\paragraph{Large Drop in Reasoning Performance of LLMs.}
% The perturbation ontology presents three distinct challenges to the LLMs: (i) demonstrating general problem-solving proficiency in the logic alteration domain; (ii) comprehending both abstract mathematical and coding concepts and principles in the problem-solving process in the \emph{concept analysis} domain;
% and (iii) maintaining robustness across various formats in the \emph{format change} and \emph{format constraint} domain. 
%\hl{[If these are related to the three highest-level categories, then these should go to the ontology section imo]}
The results show that perturbed questions significantly challenge all models in both math and coding contexts. GPT-4's accuracy decreased notably, as did other LLMs, with all showing a performance decline of over 30 points. Notably, closed-source models outperformed open-source ones in every tested aspect. Models like CodeLlama and Metamath fine-tuned on specific tasks, performed better in logic alteration and representational perturbations but worse in concept analysis. This suggests fine-tuning may restrict broader reasoning capacities. In general, LLMs handled logic alteration better than concept analysis, indicating their robustness in abstract reasoning yet limitations in understanding deeper mathematical concepts. GPT-4 demonstrated resilience across various question types, outshining others, especially in handling different problem-solving frameworks, although it still struggled more in math than in coding in concept analysis. We include Target-wise (Level II) performance analysis in Appendix~\ref{sec:fine-grained-analysis}. The latest LLMs, such as o1-preview and GPT-4o, have outperformed their predecessor, GPT-4. However, there is still a performance decline of 9\% for o1-preview and 12\% for GPT-4o. Notably, o1-preview demonstrates the highest resilience to perturbation tests in both Math and Coding questions, highlighting the significance of learning to reason.

\begin{figure*}[t]
  \centering
  \begin{subfigure}[b]{0.65\textwidth}
  \includegraphics[width=\textwidth]{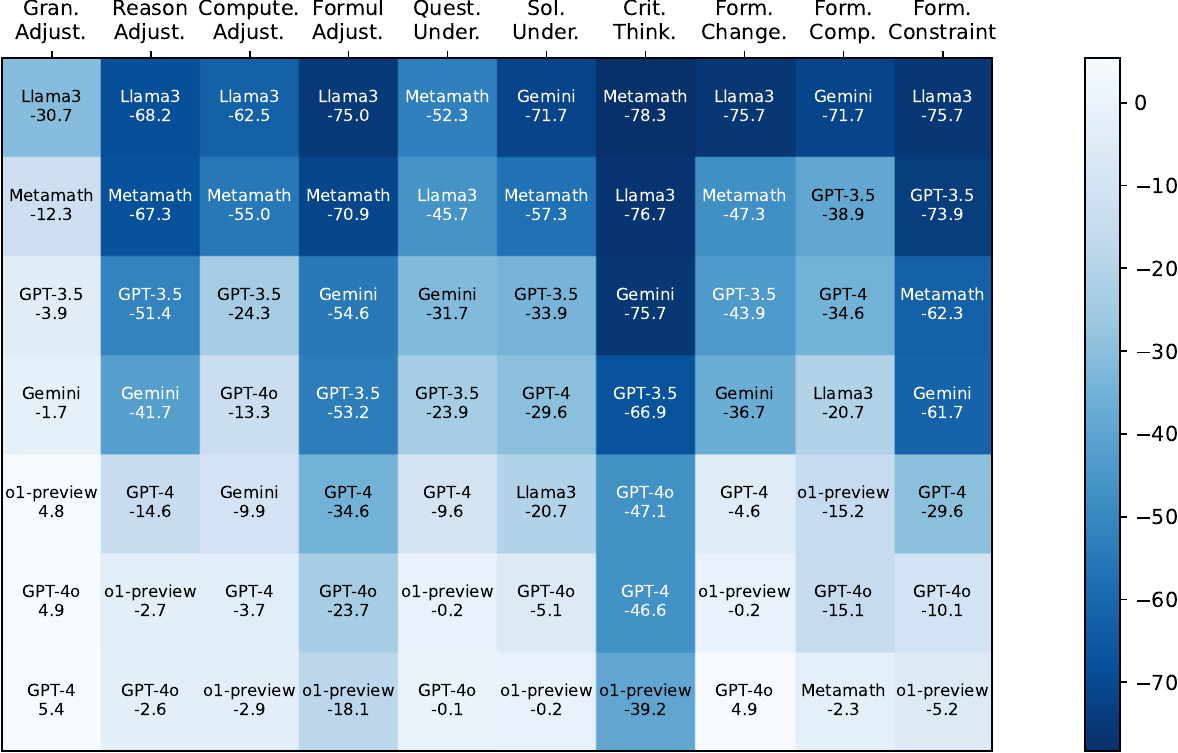}
  \caption{}
  \end{subfigure}
  \vfill
  \begin{subfigure}[b]{0.65\textwidth}
  \includegraphics[width=\textwidth]{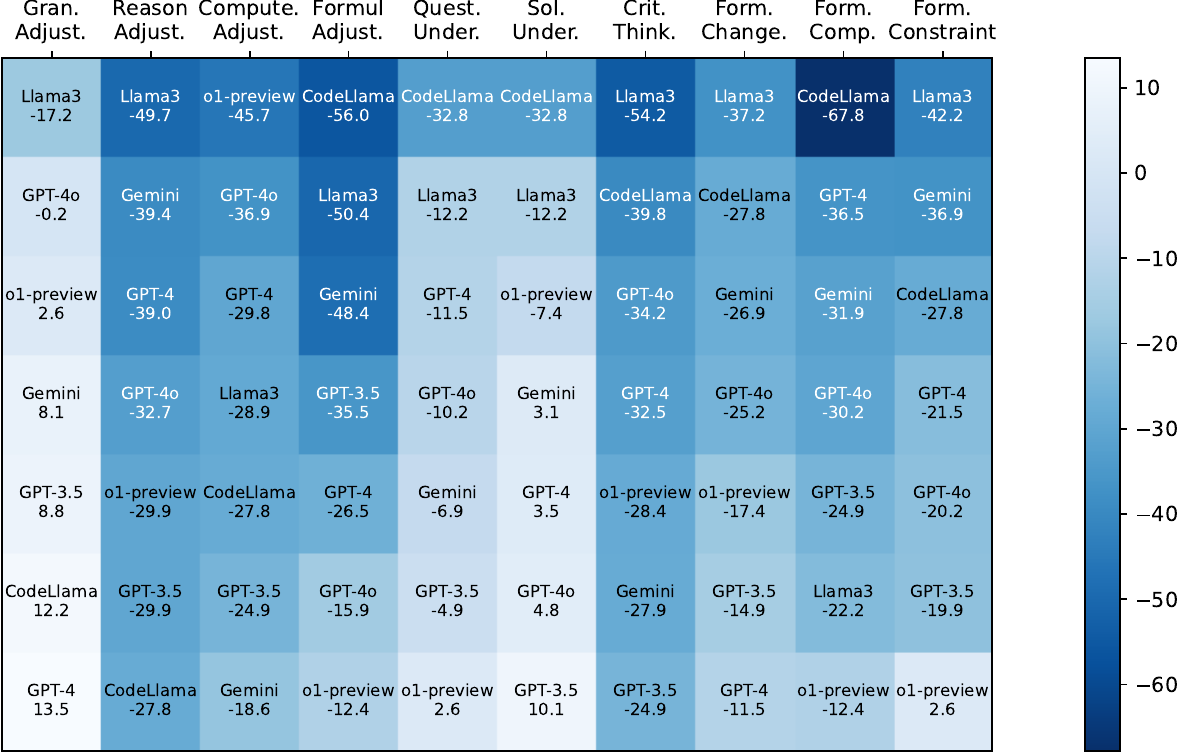}
  \caption{}
  \end{subfigure}
  \caption{Performance drop of LLMs for various dimensions (Level III) of perturbations in (a) \dataset{} and (b) \datasetcode{} as compared to the performance on the original GSM8K and HumanEval test datasets.}
  \label{fig:drop}
\end{figure*}

\subsection{Analyses} 
\label{sec:analysis}

\paragraph{Incorporation of Original Answer in Prompt.}
% In \cref{table:logic_alteration}, we investigate if providing models with the correct answer to the original question enhances their ability to solve questions with minor logical modifications (perturbations). By altering the input prompt to incorporate both the question and its gold solution, as outlined in \cref{app:exp-dets}, we observe significant improvements in model performance across various dimensions. The \emph{Computational Adjustment} dimension, especially \dataset{}, benefited the most, suggesting that models could effectively apply the original question's reasoning to tackle the altered one. In contrast, performance in the \emph{Symbolic Manipulation} dimension remained poor, indicating that access to a concrete solution does little to aid in abstract reasoning challenges. Surprisingly, even with the correct answers, Gemini and GPT-3.5 fail on simple variants of some questions. This indicates that current LLMs' low robustness to logic perturbations.

In \cref{table:logic_alteration}, providing models with the correct answer to the original question along with the prompt significantly improves their ability to solve perturbed questions, particularly in the \emph{Computational Adjustment} dimension. However, performance remains weak in \emph{Formula Adjustment}, highlighting challenges in abstract reasoning despite access to solutions. Notably, even equipped with correct answers, some models like Gemini and GPT-3.5 still fail on simpler question variants, underscoring their low sensitivity to semantic perturbations.

\paragraph{Prompting Techniques.}

% In \cref{table:logic_alteration_prompting}, we explore how different prompting techniques influence the performance in the domain of \textit{logic alteration} within \dataset{}. Specifically, we employed the Self-Consistency prompting method \citep{Wang2022SelfConsistencyIC}, and the Program of Thoughts prompting method \cite{Chen2022ProgramOT}, to gather our findings, with the experimental setup outlined in \cref{app:exp-dets}. Interestingly, we observe that the Program-of-Thoughts prompting technique significantly improved the reasoning adjustment capability across all closed-source models. This improvement is attributed to the reduction of logical errors facilitated by the use of Python programming. Additionally, we observed an 8-point performance increase in symbolic manipulation for GPT-4, suggesting that incorporating code-based chain of thought reasoning could enhance abstract reasoning abilities. However, when applying Self-Consistency prompting, only minor improvements were noted, and in the case of the Gemini model, performance declined. This indicates that this model struggles to apply the correct problem-solving procedures effectively.

In \cref{table:logic_alteration_prompting}, different prompting techniques greatly influence model performance in \emph{Logic Alteration} tasks. The Program-of-Thoughts technique notably boosts reasoning capabilities in closed-source models by reducing logical errors, leading to better performance in symbolic manipulation for GPT-4. Conversely, the Self-Consistency method shows only minor improvements and even a performance decline in the Gemini model, suggesting difficulties in effective in-context learning for new unseen tasks.

\paragraph{Identified Vulnerabilities in Reasoning.}

The \emph{Formulation Adjustment} dimension presents a significant challenge to both closed-source and open-source models, largely due to the demands of abstract reasoning. Instead of reasoning a number or code as the answer, this involves the manipulation of abstract math and coding concepts in the logical space behind the surface of the problem. For example, in the \emph{WhatIf} category, models must hypothesize outcomes by changing certain events under consistent conditions, which requires a nuanced grasp of the problem-solving framework. The \emph{Critical Thinking} dimension tests a model's ability to scrutinize relationships between pieces of information, demanding a comprehensive analysis to identify inconsistencies without a predefined solution path. This emphasizes the necessity for models to thoroughly understand and navigate through all possible avenues to effectively resolve conflicts or discrepancies. Furthermore, the \emph{Format Change} dimension poses difficulties to models like GPT-3.5 attempt to follow these constraints but often fail to maintain the integrity of their reasoning when adapting to new formats, highlighting a lack of flexibility in handling varied task demands.

\section{Discussions}

\subsection{Difficulty Change by Perturbations}

The performance drop may stem from an increased scale of reasoning or a higher level of abstract reasoning required. To explore this, experiments measured changes in the scale and depth of reasoning by comparing the number of reasoning steps and the depth required for each perturbed question against its original version. Difficulty was also evaluated through A/B testing and by recording human performance and response times across various perturbation categories as detailed in \cref{sec:difficulty_evaluation}. \cref{tab:difficulty} conducted human evaluation on 44 perturbation types, 11 increased the number of reasoning steps needed and 10 required deeper reasoning compared to original questions. Although more complex questions increased the time humans needed to respond, human performance remained almost the same. However, LLMs showed a notable decrease in performance—11.6\% for increased reasoning steps and 3.9\% for deeper reasoning. Further, There was also a more than thirty percent change in model performance for perturbed questions of equal difficulty, indicating that increased complexity have minor impacts on model performance, the major performance gap may still come from lack of robustness of LLMs.

\begin{table}[ht]
\centering
\scriptsize
\begin{NiceTabular}{l|ccc}
\toprule
\textbf{Category} & \textbf{Human Acc($\Delta$)} & \textbf{Model Acc($\Delta$)} & \textbf{Time Consumption} \\ 
\midrule
\multicolumn{4}{c}{Number of Reasoning Steps} \\ \hline
\upicon{} & 95.2(-4.8) & 32.0(-44.0) & 178\% \\ 
\equalicon{} & 98.4(-1.6) & 44.4(-31.6) & 39\% \\ 
\midrule
\multicolumn{4}{c}{Reasoning Depth} \\ \hline
\upicon{} & 97.3(-2.7) & 38.4(-37.6) & 113\% \\ 
\equalicon{} & 97.7(-2.3) & 42.1(-33.9) & 62\% \\ 
\bottomrule
\end{NiceTabular}
\caption{Summary of Human and Model Accuracy, and Time Consumption by Number of Steps and Conceptual Depth of Questions. \upicon{} indicates an increase, \equalicon{} indicates no change in reasoning steps or depth. $\Delta$ stands for the performance change relative to original question}
\label{tab:difficulty_summary}
\end{table}

\subsection{Design Choices behind Ontology}

An effective perturbation type maintains control over most variables while introducing only unidirectional changes to the original questions. Ideally, these perturbations should be noticeable to humans yet subtle enough that the required changes in skills for solving these variant questions do not significantly alter human reasoning, due to inherent human cognitive priors. Any data perturbation ontology necessitates predefined assumptions about which aspects of the data are mutable and how these changes might influence the outcomes. Therefore, recognizing and understanding these assumptions is crucial for enhancing future data augmentation efforts.  We document the aspects we have modified, the rationale behind these changes in \cref{sec:principles_behind_Ontology}.

\subsection{Scaling to More Instances}
% Our human-in-the-loop approach may hinder the effective scaling to more instances. Although scaling is not our primary objective for introducing the ontology, our focus remains on evaluating the performance across each perturbation category, independent of the perturbation difficulty, thus necessitating a human in the loop. Nevertheless, it is still possible to bootstrap an augmented dataset that can scale to more instances using a multi-agent approach. Specifically, this approach filters out the more challenging samples. Our measurements indicate that the success rate of GPT-4 in filtering out these challenging categories has reached 87\%.

Our human-in-the-loop approach may restrict scaling to more instances; however, our primary focus is on evenly evaluating performance across various perturbation categories, rather than on scaling. Nonetheless, it is feasible to expand the dataset through a multi-agent approach \citep{Wang2024BenchmarkSA}, which selectively filters out the more challenging samples. Our initial experiments, as detailed in \cref{tab:difficulty}, indicate that GPT-4 can successfully filter out challenging perturbation categories, achieving a perturbation success rate of over 90\%.

\section{Conclusion}
Our study evaluated the robustness of several prominent Large Language Models (LLMs) in handling mathematical and coding problems. By employing an ontology for random perturbations on questions from the GSM8K and HumanEval datasets, we crafted two specialized datasets, \dataset{} and \datasetcode{}, containing 216 and 219 questions respectively. These datasets target a broad variations of mathematical and coding problem-solving and analytical skills, resulting in notable performance drops in LLMs upon evaluation.
The introduction of \dataset{} and \datasetcode{} provides a new framework for assessing LLMs' abilities in mathematics and coding, while also revealing their vulnerabilities in consistent reasoning across different formats. This research highlights the complex challenges that LLMs face, stressing the importance of continued exploration into their strengths and weaknesses in logical reasoning tasks. Our dataset \dataset{}~\footnote{\url{https://huggingface.co/datasets/declare-lab/GSM8k_MORE}} 
and \datasetcode{}~\footnote{\url{https://huggingface.co/datasets/declare-lab/HumanEval_CORE}} 
will be publicly available online.

\section{Limitations}
Despite our attempt to construct a novel systematic ontology to evaluate an LM's "real" robustness and reasoning capabilities in structured reasoning tasks, it may not precisely reflect LLM's true ability due to several factors.

\paragraph{Incompleteness.}
In our endeavor to develop a comprehensive ontology for evaluating Language Models' (LMs) responses to perturbed questions across various reasoning scenarios, we recognize significant limitations. Firstly, despite our efforts, the ontology may not fully capture all essential aspects of reasoning abilities, lacking in breadth and depth. Secondly, the complexity within each reasoning category can vary significantly. For instance, within the \emph{Computation Demand} category, adjusting the number of digits in mathematical operations allows us to modulate the reasoning challenge. However, creating a benchmark that exhaustively encompasses all facets of reasoning behavior is an unattainable goal. Such an exhaustive compilation is beyond the scope of any single study and necessitates collective efforts from the broader research community.

\paragraph{Scalability.}
The size of our dataset is constrained due to the human in the loop required for its preparation. Each question generated by GPT-4 needs to be meticulously reviewed to ensure it is solvable and accurately reflects the intended perturbation specific to its category, without introducing unintended modifications. Furthermore, confirming the accuracy of answers is a critical step, as many questions do not yield answers that exactly match a predefined format. This verification process limits our ability to expand the dataset on a large scale, as it relies on manual effort.

\section{Potential Risks}
Not applicable.

\section{Ethical Considerations}

Not applicable.

\bibliography{custom}
\bibliographystyle{iclr}

\appendix

\onecolumn

\section{Recommendations}

Based on our findings, we make the following recommendations as strategies to address the weaknesses we identified in the logical reasoning competencies of LLMs.

{\bf Diversify the Datasets and Formats Used in Fine-tuning.} If a model is trained exclusively on a single problem-solving method, its capability to adapt to questions presented in different formats and solve a diverse array of problems diminishes. To counter this, we suggest boosting the model's resilience to perturbations by fine-tuning it with datasets in a variety of formats and adding augmented instructions. 

{\bf Include More Complex Open-Ended Questions.} 
It is also crucial to move beyond simple multiple-choice questions, and include open-ended questions that test the model's comprehension of mathematical concepts in the fine-tuning dataset, as this enhances its overall understanding and interpretation of questions.

\section{Dataset Details}
\label{sec:dataset_details}
In particular, there are a total of 5 math questions for each category except \emph{Change Subject} and \emph{Reverse Engineering}, which have 3 and 4 questions, respectively, in \dataset{}. Likewise, all but \emph{Reverse Engineering} perturbation---with 4 questions---have 5 coding questions in \datasetcode{}.

\section{Experiment Details}
\subsection{Model Details}
\label{sec:model_details}
\begin{itemize}
    \item We use o1-preview with model version dated "2024-08-06".
    \item We use GPT-4o with model version dated "2024-08-06".
    \item We use GPT-4 with model version named "1106-Preview".
    \item For GPT-3.5, we use version "GPT-3.5-turbo-0613".
    \item We use Gemini-1.5-pro for Gemini model.
    \item Llama3-Instruct \url{https://huggingface.co/meta-llama/Meta-Llama-3-8B-Instruct}.
    \item MetaMath-70B-V1.0 \url{https://huggingface.co/meta-math/MetaMath-70B-V1.0}.
    \item CodeLlama-70B-Instruct \url{https://huggingface.co/codellama/CodeLlama-70b-Instruct-hf}.
\end{itemize}

\subsection{Filtering Criteria}
\label{sec:properties}
\begin{enumerate}[label=(\roman*), leftmargin=*, wide, labelwidth=0pt, labelindent=0pt, noitemsep]
    \item \textbf{Human Understandability}: The generated questions should be comprehensible to humans. The language, structure, and presentation of the questions should be clear and easy to understand. Vague or confusing questions should be rejected.
    \item \textbf{Logical Coherence}: The questions must make logical sense. They should not contain contradictions\footnote{Except for the \textit{conflicting information} (\ref{gs:conflicting-information}) type, where we intentionally introduce contradictions.}, nonsensical premises, or incoherent elements.
    \item \textbf{Instruction Adherence}: The generated questions should closely adhere to the instructions in the prompt for the specific perturbation type. 
    % This includes following the guidelines of the template used for generating these questions. 
    The question should not deviate from the intended method of perturbation. 
\end{enumerate}

\section{Fine-Grained Analysis}
\label{sec:fine-grained-analysis}
As illustrated in \cref{table:main_result}, the introduction of perturbed questions poses significant challenges to all models in both math and coding contexts. Specifically, GPT-4's accuracy decreased from 100\% to 74.2\% and from 80\% to 56.7\% in math and coding scenarios respectively. This trend of performance degradation is even more pronounced in other LLMs, with all experiencing a decline exceeding 30 points in their weighted average performance across both the mathematical and coding datasets. For instance, GPT-3.5 witnessed a dramatic performance reduction from 80\% to 35.75\% on the mathematical dataset and from 80\% to 47.09\% for the coding dataset.

Notably, closed-source models consistently outperform open-source models in every tested dimension. Additionally, it has been observed that models which have undergone fine-tuning on task-specific data---such as, CodeLlama for coding problems and Metamath for math problems---show enhanced performance in the areas of \emph{logic alteration} and \emph{representational perturbations} as compared to the Llama2-Chat model. However, this fine-tuning process appears to compromise Llama2's capabilities within the \emph{concept analysis} domain. This observation suggests that the focus of fine-tuned, task-specific data on deriving a fixed solution might limit a model's broader capacity for reasoning, thereby affecting its ability to analyze and comprehend the underlying problem-solving process.

\noindent
\textbf{(Level II) Target-wise Performance.} Following \cref{table:main_result}, LLMs generally showed better results on \emph{logic alteration} questions, which involve concrete reasoning steps in problem-solving. Despite this, even the state-of-the-art models struggled with certain perturbed versions of these questions. This indicates that while current models may possess general task-solving skills and abstract reasoning ability, there is still a limitation in their reasoning robustness when faced with altered logic. On the other hand, \emph{concept analysis} questions, which demand a deeper understanding of mathematical concepts and problem-solving frameworks, resulted in lower success rates. This suggests that while current models can find correct answers, they may lack a systematic logical framework for problem-solving and struggle with analyzing and understanding different concepts necessary to answer the question.

GPT-4, in particular, demonstrated superior performance across all categories, showing increased resilience to changes in question format and expected responses. This contrasts with other models, which performed poorly on tasks involving representational perturbations, hinting at a limitation in transferring their reasoning processes to different formats. Interestingly, the average performance decline across domains was similar for both math and coding contexts, with the notable exception of the \emph{concept analysis} domain, where the drop in math performance was 21\% greater than in coding. This discrepancy suggests that  LLMs may possess a more profound understanding of problem-solving frameworks in coding contexts compared to mathematical ones.

\section{Benchmark Difficulty Evaluation}
\label{sec:difficulty_evaluation}
The evaluation of difficulty was conducted by three undergraduate students. Each participant was presented with questions to solve on paper, without access to calculators or computers. Their task completion time for each question was recorded. The students also documented changes in the number of steps required to solve perturbed questions compared to the original, noting whether the number of steps increased, or remained roughly the same. Additionally, they assessed whether the perturbed variants demanded more higher level mathematical concepts or skills.

\begin{table*}[h!]
\centering
\resizebox{\textwidth}{!}{%
\begin{tabular}{l|l|ccccc}
\toprule
\textbf{Dimension} & \textbf{Category} & \textbf{Human Acc} & \textbf{Model Acc} & \textbf{Time Consump} & \textbf{Steps} & \textbf{Reasoning Depth}  \\
\midrule
\multirow{4}{*}{Granularity Adjustment} & Remove Constraint & 100 & 84 & -70 & \equalicon & \equalicon \\ 
& Partial Solution & 100 & 70 & -40 & \equalicon & \equalicon \\ 
& Solution Plan & 100 & 76 & -50 & \equalicon & \equalicon \\ 
& Detail Expansion & 100 & 70 & -50 & \equalicon & \equalicon\\ 
\midrule
\multirow{8}{*}{Reasoning Adjustment} & Add Restriction & 100 & 22 & +100 & \upicon & \equalicon\\ 
& Subsequent Question & 100 & 34 & +50 & \equalicon & \equalicon\\ 
& Concurrent Question & 100 & 36 & +150 & \equalicon & \equalicon \\ 
& Change Question & 100 & 42 & -70 & \equalicon & \upicon \\ 
& Info Recombination & 87 & 28 & +40 & \upicon & \equalicon \\ 
& Domain Knowledge & 80 & 56 & +450 & \upicon & \upicon \\ 
& Complex Reality & 100 & 32 & +100 & \upicon & \equalicon \\ 
& General Solution & 100 & 24 & +0 & \equalicon & \upicon\\
\midrule
\multirow{3}{*}{Computation Adjustment} & Computation Demand & 100 & 36 & +20 & \equalicon & \equalicon \\ 
& Change Value & 100 & 56 & -10 & \equalicon & \equalicon\\ 
& Change Operation & 100 & 66 & +0 & \equalicon & \equalicon\\
\midrule
\multirow{7}{*}{Formulation Adjustment} & Symbolic Response & 100 & 42 & +100 & \equalicon & \upicon\\ 
& Value Relationship & 93 & 20 & +100 & \equalicon & \upicon\\ 
& Variable Group & 100 & 24 & +140 & \upicon & \upicon\\ 
& Backward Reasoning & 100 & 26 & +100 & \equalicon & \upicon\\ 
& Counterfactual & 100 & 18 & +160 & \upicon & \equalicon \\ 
& Solve Value & 100 & 28 & +140 & \equalicon & \equalicon\\ 
& Identify Range & 93 & 26 & -40 & \equalicon & \equalicon\\
\midrule
\multirow{4}{*}{Question Understanding}& Inherent Premise & 100 & 38 & +160 & \equalicon & \equalicon\\ 
& Complete Missing & 100 & 60 & -50 & \equalicon & \equalicon\\ 
& Question Formulation & 93 & 50 & +200 & \equalicon & \equalicon\\ 
& Add Misinformation & 100 & 68 & +50 & \equalicon & \equalicon\\
\midrule
\multirow{4}{*}{Solution Understanding}& Optimize Solution & 100 & 50 & +160 & \upicon & \upicon\\ 
& Step Functionality & 100 & 42 & +100 & \equalicon & \upicon\\ 
& Theoretical Basis & 100 & 62 & -50 & \equalicon & \equalicon\\ 
& Cost Analysis & 100 & 58 & +50 & \equalicon & \upicon\\ 
\midrule
\multirow{5}{*}{Critical Thinking}& Seek Clarification & 80 & 26 & -50 & \equalicon & \equalicon\\ 
& Conditional Analysis & 93 & 16 & +200 & \upicon & \equalicon\\ 
& Conflicting Information & 100 & 8 & +50 & \equalicon & \equalicon\\ 
& Surface Error & 100 & 44 & +50 & \equalicon & \equalicon\\ 
& Hidden Error & 93 & 30 & +200 & \equalicon & \equalicon \\ 
\midrule
\multirow{4}{*}{Format Change} & Setting Rephrase & 100 & 50 & +0 & \equalicon & \equalicon \\ 
& Change Sequence & 100 & 52 & +0 & \equalicon & \equalicon\\ 
& Close Format & 93 & 36 & +20 & \equalicon & \equalicon\\ 
& Data Restructuring & 100 & 40 & +160 & \upicon & \equalicon\\ 
\midrule
\multirow{1}{*}{Format Comparison} & Identical Problem & 87 & 42 & +20 & \equalicon & \equalicon\\ 
\midrule
\multirow{4}{*}{Format Constraint} & Reasoning Format & 100 & 30 & +200 & \upicon & \equalicon\\ 
& Reasoning Style & 100 & 34 & +170 & \equalicon & \equalicon \\
& Alternative Answer & 100 & 28 & +60 & \equalicon & \equalicon \\
& New Rule & 87 & 36 & +250 & \upicon & \equalicon \\
\midrule
Average & & 97.7 & 41.2 & +74.3 & N/A & N/A  \\
\bottomrule
\end{tabular}
}
\caption{Comparison of Average Baselines: Human vs. Models. Displays accuracy rates for participants and models, and time change percentage for solving perturbed vs. original questions. \upicon{} indicates an increase; \equalicon{} signifies equal reasoning depth.}
\label{tab:difficulty}
\end{table*}

% \section{Auto Evaluation Results}

% We also use GPT-4o to automatically evaluate the outputs generated by the models for faster assessment. However, we observed that the auto-evaluated results tend to show lower performance, likely because the generated answers can take multiple formats, while we only provide the simplest form in our reference answers. Despite this, the results from auto-evaluation are not significantly different from those of manual evaluation.

\section{Original Questions from GSM8K}
\label{app:original_gsm8k}

The following selected questions are from the GSM8K dataset, specifically chosen for their variations in complexity. Each of the five questions requires between 3 to 7 steps to solve, illustrating the range of reasoning complexity present in the GSM8K dataset. These questions span a wide array of everyday topics that involve the application of mathematical principles, including physical dimensions, profit maximization, purchasing decisions, time management, and solving multi-variable equations. Those 5 questions demands diversity of mathematical problem-solving skills in different situations.

\subsection{Question 1}
\label{math-example:merchant}
\begin{mdframed}[backgroundcolor=pink!20] 
    A merchant wants to make a choice of purchase between 2 purchase plans: jewelry worth \$5,000 or electronic gadgets worth \$8,000. His financial advisor speculates that the jewelry market will go up 2.5\% while the electronic gadgets market will rise 1.2\% within the same month. If the merchant is looking to maximize profit at the end of this month by making a choice, how much profit would this be?\\
    
    \textbf{Answer:} If he purchases jewelry, he will make a profit of 2.5\% which is 5000*(2.5/100) = 125.  If he purchases electronic gadgets, he will make a profit of 1.2\% which is 8000*(1.2/100) = 96. If he wants to maximize profit, since 125 > 96, he will choose to purchase jewelry, thereby making a profit of 125\\
    
\end{mdframed}

\subsection{Question 2}
\label{math-example:john}
\begin{mdframed}[backgroundcolor=pink!20] 
    \textbf{Question 2:} John has 3 boxes. Each box is 5 inches by 6 inches by 4 inches. The walls are 1 inch thick. What is the total inner volume of all 3 boxes?\\
    
    \textbf{Answer:} The walls subtract 2*1=2 inches from each dimension. So each box has 5-2=3 inch width It also has a 6-2=4 inch height. Finally, it has a 4-2=2 inch depth. So the inner volume of one box is 4*3*2=24 cubic inches. So in total the inner volume of the 3 boxes is 3*24=72 cubic inches\\
\end{mdframed}

\subsection{Question 3}
\label{math-example:kylar}
\begin{mdframed}[backgroundcolor=pink!20] 
    \textbf{Question 3:} Kylar went to the store to buy glasses for his new apartment. One glass costs \$5, but every second glass costs only 60\% of the price. Kylar wants to buy 16 glasses.  How much does he need to pay for them?\\
    
    \textbf{Answer:} The discount price of one glass is 60/100 * 5=3.  If every second glass is cheaper, that means Kylar is going to buy 16 / 2 = 8 cheaper glasses.  So for the cheaper glasses, Kylar is going to pay 8 * 3 = 24. And for the regular-priced glasses, Kylar will pay 8 * 5 = 40.  So in total Kylar needs to pay 24 + 40 = 64 for the glasses he wants to buy.\\
    
\end{mdframed}

\subsection{Question 4}
\label{math-example:vicki}
\begin{mdframed}[backgroundcolor=pink!20] 
    \textbf{Question 4:} Vicki is planning a pop concert at her high school. The show will be 2 hours. She is allowing each group 2 minutes to get on stage, 6 minutes to perform, and then 2 minutes to exit the stage. If she allows a 10-minute intermission, how many groups can perform in the concert?\\
    
    \textbf{Answer:} First, we should convert the 2 hours of showtime into minutes for our calculations. Since there are 60 minutes in 1 hour, the show will be 2 x 60 = 120 minutes.  Of those 120 minutes, 10 will be used for intermission, so 120 – 10 = 110 minutes for performances. Each group will use 2 minutes to get on stage + 6 minutes to perform + 2 minutes to exit the stage = 10 minutes of show time. Of the 110 minutes of performances, 10 are used per group, so 110 minutes / 10 = 11 groups can perform.\\
    
\end{mdframed}

\subsection{Question 5}
\label{math-example:lily}
\begin{mdframed}[backgroundcolor=pink!20] 
    \textbf{Question 5:} Together Lily, David, and Bodhi collected 43 insects. Lily found 7 more than David. David found half of what Bodhi found.  How many insects did Lily find?\\
    
    \textbf{Answer:} Let B = the number of insects Bodhi collected. David = B/2, Lily = B/2 + 7. B + B + 7 = 43. Simplify: 2B = 36. Simplify B = 18 insects. David = 18/2 =9 insects. Lily = 9 + 7 = 16 insects. Lily found 16 insects.
\end{mdframed}

\section{Original Questions from HumanEval}
\label{app:original_humaneval}
The following selected questions are from the HumanEval dataset, specifically chosen for their variations in complexity. Each of the five questions requires different number of lines code to solve, illustrating the range of reasoning complexity present in the HumanEval dataset. These questions includes basic programming concepts such as string manipulation, list indexing, classic algorithm, math problem and state conditions. Those 5 questions demands diversity of programming skills and concepts in different situations.
\definecolor{codegreen}{rgb}{0,0.6,0}
\definecolor{codegray}{rgb}{0.5,0.5,0.5}
\definecolor{codepurple}{rgb}{0.58,0,0.82}
\definecolor{codebg}{rgb}{0.95,0.95,0.92}

\lstdefinestyle{mystyle}{
    backgroundcolor=\color{codebg},   
    commentstyle=\color{codegreen},
    keywordstyle=\color{magenta},
    numberstyle=\tiny\color{codegray},
    stringstyle=\color{codepurple},
    basicstyle=\ttfamily\footnotesize,
    breakatwhitespace=false,         
    breaklines=true,                 
    captionpos=b,                    
    keepspaces=true,                 
    numbers=left,                    
    numbersep=5pt,                  
    showspaces=false,                
    showstringspaces=false,
    showtabs=false,                  
    tabsize=2
}
\subsection{Question 1}
\label{code-example:flip-case}
\begin{lstlisting}[language=Python, style=mystyle]
def flip_case(string: str) -> str:

    """For a given string, flip lowercase characters to uppercase and uppercase to lowercase.

    >>> flip_case('Hello')
    'hELLO'
    """
    return string.swapcase()
\end{lstlisting}

\subsection{Question 2}
\label{code-example:gcd}
\begin{lstlisting}[language=Python, style=mystyle]
def greatest_common_divisor(a: int, b: int) -> int:

    """ Return a greatest common divisor of two integers a and b

    >>> greatest_common_divisor(3, 5)
    1
    >>> greatest_common_divisor(25, 15)
    5
    """

    while b:
        a, b = b, a \% b
    return abs(a)
\end{lstlisting}
\subsection{Question 3}
\label{code-example:derivative}
\begin{lstlisting}[language=Python, style=mystyle]
def derivative(xs: list):

    """ xs represent coefficients of a polynomial.
    xs[0] + xs[1] * x + xs[2] * x^2 + ....
    Return derivative of this polynomial in the same form.

    >>> derivative([3, 1, 2, 4, 5])
    [1, 4, 12, 20]
    >>> derivative([1, 2, 3])
    [2, 6]
    """
    if len(xs) == 1: return [0]
    if len(xs) == 0: return []
    return [(i * x) for i, x in enumerate(xs)][1:]
\end{lstlisting}
\subsection{Question 4}
\label{code-example:sum-squares}
\begin{lstlisting}[language=Python, style=mystyle]
def sum_squares(lst):

    """
    This function will take a list of integers. For all entries in the list, the function shall square the integer entry if its index is a 
    multiple of 3 and will cube the integer entry if its index is a multiple of 4 and not a multiple of 3. The function will not 
    change the entries in the list whose indexes are not a multiple of 3 or 4. The function shall then return the sum of all entries. 

    Examples:
    For lst = [1,2,3] the output should be 6
    For lst = []  the output should be 0
    For lst = [-1,-5,2,-1,-5]  the output should be -126
    """
    
    result =[]
    for i in range(len(lst)):
        if i%3 == 0:
            result.append(lst[i]**2)
        elif i% 4 == 0 and i%3 != 0:
            result.append(lst[i]**3)
        else:
            result.append(lst[i])
    return sum(result)
\end{lstlisting}
\subsection{Question 5}
\label{code-example:is-nested}
\begin{lstlisting}[language=Python, style=mystyle]
def is_nested(string):

    """
    Create a function that takes a string as input which contains only square brackets.
    The function should return True if and only if there is a valid subsequence of brackets 
    where at least one bracket in the subsequence is nested.
    Examples:
    [[]] output: True
    [][] output: False
    [] output: False
    [[][]] output: True
    [[]][[ output: True
    """

    stack = []
    depth = 0
    for i, char in enumerate(string):
        if char == '[':
            stack.append('[')
            if depth > 0:
                depth -= 1
        elif char == ']':
            if len(stack) > 0:
                stack.pop()
                depth += 1
            if depth >= 2:
                return True
            if len(stack) == 0:
                depth = 0
    return False
\end{lstlisting}

\section{Ontology}
The summary of our proposed ontological categories is shown in \Cref{tab:new_ontology}.

\section{Ontology of Perturbations}
\label{sec:ontology}
% \begin{figure*}[ht!]
%   \centering
%   \includegraphics[width=\textwidth]{latex/images/ontology_uni.dot.pdf}
%   \caption{The ontology of the perturbations.
%   % The \textcolor{red}{red} dashed edges represent \emph{enhanced} to \emph{primary} relationships.
%   }
%   \label{fig:ontology}
% \end{figure*}

\subsection{Principles behind Ontology}
\label{sec:principles_behind_Ontology}
Consider a maths question:
\begin{mdframed}[backgroundcolor=yellow!20]
\textbf{Question:} John has 3 boxes, each of which is externally measured as 5 inches by 6 inches by 4 inches. The boxes have walls that are 1 inch thick. What is the total inner volume of all the boxes?
\end{mdframed}
We consider the following eight aspects of such questions:
\begin{enumerate}[label=(\roman*), leftmargin=*, wide, labelwidth=0pt, labelindent=0pt]
\item \textbf{Information}: Each sentence clause that is mentioned inside the question. For example: \texttt{Each box is 5 inches by 6 inches by 4 inches.} 

\item \textbf{Query}: What is being asked by the Question that can be calculated with the given Information? For example: \texttt{What is the total inner volume of all the boxes?}

\item \textbf{Values:} Values inside the Information. For example, \texttt{3} boxes in this particular instance.

\item \textbf{ToolBox}: Mathematical concepts, formulas, and operations that are relevant to solving a specific problem. For example: \texttt{Multiplication is used to calculate the volume of a rectangular prism (box) as length × width × height} and \texttt{Subtraction is used to adjust the external dimensions to account for the wall thickness.}

\item \textbf{Mathematical Structure}: Chain of thought and problem-solving strategies that outline how the `Tools' in Toolbox are organized to transition from the given data to the final answer. For example, to solve the question above:  \texttt{first, Subtract the thickness of the walls; second, calculate the volume of one box; third, multiply the volume of one box by the number of boxes. }

\item \textbf{Query Representation}: The Format of how Information and Values are presented. For example: the sequence of Information presented.

\item \textbf{Final Answer}: Final answer to the Query. For example: \texttt{72}

\item \textbf{Answer Representation}: The Format of the answer presented.

\end{enumerate}

In a similar vein, consider a coding question:
\definecolor{codegreen}{rgb}{0,0.6,0}
\definecolor{codegray}{rgb}{0.5,0.5,0.5}
\definecolor{codepurple}{rgb}{0.58,0,0.82}
\definecolor{codebg}{rgb}{0.95,0.95,0.92}

\lstdefinestyle{mystyle}{
    backgroundcolor=\color{codebg},   
    commentstyle=\color{codegreen},
    keywordstyle=\color{magenta},
    numberstyle=\tiny\color{codegray},
    stringstyle=\color{codepurple},
    basicstyle=\ttfamily\footnotesize,
    breakatwhitespace=false,         
    breaklines=true,                 
    captionpos=b,                    
    keepspaces=true,                 
    numbers=left,                    
    numbersep=5pt,                  
    showspaces=false,                
    showstringspaces=false,
    showtabs=false,                  
    tabsize=2
}
\begin{lstlisting}[language=Python, style=mystyle]
def greatest_common_divisor(a: int, b: int) -> int: 
    """ Return the greatest common divisor of two integers a and b 
    Example:
    >>> greatest_common_divisor(3, 5) 
    1 
    >>> greatest_common_divisor(25, 15) 
    5 
    """
\end{lstlisting}
We can decompose the coding question into the following aspects:
\begin{enumerate}[label=(\roman*), leftmargin=*, wide, labelwidth=0pt, labelindent=0pt]
    \item \textbf{Question Header}: The name of the function, in the case above, \texttt{\footnotesize greatest\_common\_divisor}  \\
    \item \textbf{Docstring}: Defines the requirement for the final output. For example, \texttt{\footnotesize Return the greatest common divisor of two integers}
    \item \textbf{Values}: The type and structure of input arguments. In the above example, \texttt{\footnotesize a (integer type)} and \texttt{\footnotesize b (integer type)}
    \item \textbf{Examples}: Demonstrations of how the function is used. In the case above,
    \begin{lstlisting}[language=Python, style=mystyle]
"""
>>> greatest\_common\_divisor(3, 5) 
1 
>>> greatest\_common\_divisor(25, 15) 
5 
"""\end{lstlisting}
        \item \textbf{Toolbox}: Libraries and operations that can be used to achieve a function.
        \item \textbf{Code Structure}: Sequence of steps of code to fulfill the requirement specified in \textbf{Docstring}
        \item \textbf{Question Representation}: Format of how the \textbf{Question header} and \textbf{Docstring} is presented
        \item \textbf{Answer Representation}: Format of how the \textbf{Code Structure} is presented.
\end{enumerate}

The perturbations in the ontology we introduce operate on these eight aspects of a maths or coding question. Each perturbation changes only one or two aspects of the original question.
%Our ontology focuses on three fundamental aspects of Large Language Models (LLMs): \hl{Problem Solving, Instruction Following, and Robustness}. Each of these aspects is further divided into multiple dimensions. In particular, for the domains of mathematics and coding, we apply specific methods to introduce variations or \emph{perturbations} to the questions along these dimensions. While these methods are uniquely tailored to coding and mathematics, they share underlying similarities. Each dimension can manifest in various ways that correspond to some method of perturbation. These perturbations can affect the difficulty level of the questions, making them either more challenging or simpler. Additionally, some perturbations may result in questions that do not have a definitive answer.

We broadly group these perturbations into two main categories: \emph{Structural Perturbation} and \emph{Representational Perturbation}. \emph{Structural Perturbations} generate new questions by modifying the specific targeted aspects of inherent logic, framework, or concepts in the original question. \emph{Structural Perturbation} is further categorized into \emph{Logic Alteration} and \emph{Concept Analysis}. \emph{Logic-Alteration} perturbations changes the logic underpinning a problem through addition or removal of information, or it changes the reasoning framework of the original problem. The \emph{Concept Analysis} questions, however, examines the underlying concepts and principles of the problem. Rather than solving a specific problem, these questions focus on analyzing the process of problem solving, and how it get the solutions, which may require a deeper understanding of the question and problem solving framework. Details and examples for each of these perturbation types are presented below. 

Unlike \emph{Structural Perturbations}, \emph{Representational Perturbations} retain the logical structure of the original solution, only to exclusively change the representation or encoding of the information present in the question or in the answer. In our ontology, \emph{Representational Perturbation} has only two manifestations, \emph{Format Change}, which directly alters the representation of the questions and answers. \emph{Format Constraint}, which add constraint that indirectly alters the format of the answers. More details and examples are below.

For each of the above broad perturbation types, we further define many dimensions of perturbations. We apply specific methods to introduce variations or \emph{perturbations} to the questions along these dimensions. Each dimension can further manifest in various ways that correspond to some method of perturbation.  For example, a dimension such as ``simplify question'' can be realized in different ways for the ``logic alteration'' perturbation type.
These perturbations can affect the difficulty level of the questions, making them either more challenging or simpler. Additionally, some perturbations may result in questions that do not have a definitive answer. 

% \hl{The problem decomposition should come here, I think. It should also be named better: `problem aspects` perhaps. Also, mention how some aspects are exclusively pertinent to certain categories}

\subsection{Logic Alteration}
\label{sec:closed-question}
% Here we group perturbation operations to closed questions with definite answers, which results in a new  closed question. Theoretically, a human can come up with perturbations that transition from closed to open or vice-versa. However, performing such transitions automatically in a controlled manner and evaluating them is hard. Therefore, the original and the perturbed type of question remains same for us. Among closed questions, we list three variety of perturbations: 1) simplify questions that simplifies a question without making it unsolvable, 2) change the question such that underlying mathematical structure is changed, 3) change the value(s) in the original question.

This category groups all the perturbations that have a definitive final answer. The final answer can be in the format of a value (Math) or code (HumanEval) (for dimension ``Question Simplification'', ``Reasoning Adjustment'', ``Computation Adjustment'') or a mathematical expression (Math) or Natural Language (Code) (for dimension ``Symbolic Reasoning''). For logic alteration questions, if the final answer is normalized to the most simplified form. The generated answer can be deemed correct only if it can also normalize to the same form.

\begin{enumerate}[label=(\roman*), leftmargin=*, wide, labelwidth=0pt, labelindent=0pt]
    \item \textbf{Granularity Adjustment}: This dimension aims to make the question easier to solve. It can achieve this by using four ways:
        \begin{enumerate}[itemsep=0pt, leftmargin=*,  labelwidth=0pt, labelindent=0pt, parsep=0pt, label=\textbf{G\arabic*}.,resume]
        
\item \emph{Remove Constraint}: Remove one piece of constraint that make the question easier to solve\\
\emph{Remove Constraint (Math):} Delete one piece of \textbf{information} from the original question that does not make the question unsolvable. The aim is to simplify the question. Example:\\
\textbf{Changed from \ref{math-example:john}: } 
\begin{mdframed}[backgroundcolor=pink!20] 
John has 3 boxes. Each box is 5 inches by 6 inches by 4 inches. What is the total volume of all 3 boxes?
\end{mdframed}
\emph{Remove Constraint (Code):} Simplify the coding requirement by removing one constraint or transformation in the \textbf{Docstring} \\
Generate a python function that fulfills the requirement in docstring and examples usages below.\\
\textbf{Changed from \ref{code-example:flip-case}:}
\begin{lstlisting}[language=Python, style=mystyle]
def change_case(string: str) -> str:

    """For a given string, convert all uppercase characters to lowercase.

    >>> change_case('Hello')
    'hello'
    """
\end{lstlisting}
\label{gs:remove-constraint}

\item \emph{Partial Solution}: The answer only need to solve parts of the original question \\
\emph{Median Inquiry}: Change the original \textbf{query} to ask one of the intermediate values that is used to solve for the final answer of the original query. The aim is to simplify the question. Example:\\
\textbf{Changed from \ref{math-example:john}}: 
\begin{mdframed}[backgroundcolor=pink!20] 
John has 3 boxes. Each box is 5 inches by 6 inches by 4 inches. The walls are 1 inch thick. What is the inner volume of one box? 
\end{mdframed}
\emph{Helper Function}: Provide a helper function alongside the coding question that achieves partial function in \textbf{Code Structure} \\
\textbf{Changed from \cref{code-example:flip-case}:}
Generate a python function that fulfills the requirement in docstring and examples usages below. You should complete the function using helper function.
\begin{lstlisting}[language=Python, style=mystyle]
def helper_function(char: str) -> str:
    """Checks if a given character is uppercase or lowercase, and flips its case."""

    if char.isupper():
        return char.lower()
    elif char.islower():
        return char.upper()
    else:
        return char

def flip_case(string: str) -> str:

    """For a given string, flip lowercase characters to uppercase and uppercase to lowercase by using the helper function above to achieve the requirement
    >>> flip_case('Hello')
    'hELLO'
    """

"""
\end{lstlisting}

\label{gs:partial-solution}

\item \emph{Solution Plan}: Besides the original question, provide a high level plan of how the question should be answered, the solution will only need to execute the abstract plan.\\
\emph{Solution Plan (Math)}: Provide the original question along with its \textbf{mathematical structure} (problem strategy) to the question, ask the model to solve the question by following the strategy. \\
\textbf{Changed from \ref{math-example:john}:} 
\begin{mdframed}[backgroundcolor=pink!20] 
John has 3 boxes. Each box is 5 inches by 6 inches by 4 inches. The walls are 1 inch thick. What is the total inner volume of all 3 boxes? Follow this plan to solve the question:
[\#Solution Plan\#]
Subtract the thickness of the walls from each dimension of the box to get the inner dimensions.
Determine the width, height, and depth of the inner box.
Calculate the inner volume of one box by multiplying the width, height, and depth.
Calculate the total inner volume by multiplying the inner volume of one box by the number of boxes.
\end{mdframed}
\emph{Solution Plan (Code)}: Provide the high level plan of how the code need to be written along with the question. 
\textbf{Changed from \ref{code-example:flip-case}:}
Generate a python function that fulfills the requirement in docstring and examples usages below. You should follow the solution plan when solving the problem.
\begin{lstlisting}[language=Python, style=mystyle]
def flip_case(string: str) -> str:
    """
    Inverts the case of each character in the provided string.

    This function takes a string as an argument and returns a new string with each character's case inverted. 
    Uppercase letters are converted to lowercase, and lowercase letters are converted to uppercase.

    Solution Plan:
    1. Create a result variable to hold the updated string.
    2. Iterate through each character in the string.
    3. Check if the character is uppercase; if so, convert it to lowercase and add it to the result.
    4. If the character is lowercase, convert it to uppercase and add it to the result.
    5. After iterating through all characters, return the result.
    """
\end{lstlisting}
\label{gs:solution-plan}

\item \emph{Detail Expansion}: Besides the original question, provide a few key important details or explanations without which is hard to solve the question.\\
\emph{Detail Elaboration}: Provide original question along with the \textbf{toolbox} (commonsense knowledge) to solve the question.  \\
\textbf{Changed from \ref{math-example:john}:} 
\begin{mdframed}[backgroundcolor=pink!20] 
John has 3 boxes. Each box has outside dimensions of 5 inches by 6 inches by 4 inches. The walls of each box are 1 inch thick, uniformly throughout each face of the box, thereby reducing the inner dimensions of each box. The material of the boxes is uniformly distributed and does not bulge or cave in thereby affecting the inner volume. There are no internal structures or partitions inside the boxes that could further reduce the inner volume. What is the total inner volume of all 3 boxes?
\end{mdframed}
\emph{Example Detail:} Besides providing the input and output of each \textbf{example}, it also provide a step by step explanation of how the input is transformed to the output.
\textbf{Changed from \ref{code-example:derivative}:}
Generate a python function that fulfills the requirement in docstring and examples usages below.
\begin{lstlisting}[language=Python, style=mystyle]
def derivative(xs: list):
    """ xs represent coefficients of a polynomial.
    xs[0] + xs[1] * x + xs[2] * x^2 + ....
    Return derivative of this polynomial in the same form.

    >>> derivative([3, 1, 2, 4, 5]) calculates the derivative as [1*1, 2*2, 3*4, 4*5] resulting in [1, 4, 12, 20].

    >>> derivative([1, 2, 3]) calculates the derivative as [1*2, 2*3] resulting in [2, 6].
    """
\end{lstlisting}
\label{gs:detail-expansion}
\end{enumerate}

\item \textbf{Reasoning Adjustment}: This dimension targets to partially change the logical structure of the original problem. It can be achieved through eight ways:

\begin{enumerate}[itemsep=0pt, leftmargin=*,  labelwidth=0pt, labelindent=0pt, parsep=0pt, label=\textbf{G\arabic*}., resume]

\item \emph{Add Restriction}: Add a new piece of condition or requirement to the answer of the question. \\
\emph{Restrict Question}: Adding a new piece of \textbf{information} that serves as a constraint or modifier on the query. Example:\\
\textbf{Changed from \ref{math-example:john}}: 
\begin{mdframed}[backgroundcolor=pink!20] 
John has 3 boxes, each of which is externally measured as 5 inches by 6 inches by 4 inches. The boxes have walls that are 1 inch thick. There is also an added wooden board divider in the middle across the smallest dimension which is also 1 inch thick. What is the total inner volume of all the boxes?
\end{mdframed}
\emph{Restrict Requirement}: Add a piece of information that serves as a constraint or modifier on the function. \\
\textbf{Changed from \ref{code-example:flip-case}}
\begin{lstlisting}[language=Python, style=mystyle]
def flip_case(string: str, index: int) -> str:

    """For a given string, flip lowercase characters to uppercase and uppercase to lowercase. Only flip the case for characters at indices which are multiples of the provided index.
    Note: If the index provided is 2, only the characters at the 2nd, 4th, 6th positions and so on will have their cases flipped.
    
    >>> flip_case('Hello', 2)
    'HeLlO'
    """
\end{lstlisting}
\label{gs:add-restriction}

\item \emph{Subsequent Question}: Adding an additional query or requirement based on the answer of of the original question. \\
\emph{Further Question}: Adding an additional \textbf{query} that will need extra steps of calculation based on the final answer of the original query. 
\begin{mdframed}[backgroundcolor=pink!20] 
\textbf{Changed from \ref{math-example:john}}: John has 3 boxes. Each box is 5 inches by 6 inches by 4 inches. The walls are 1 inch thick. What is the total inner volume of all 3 boxes? \textbf{If John wants to entirely fill these boxes with small cubes each measuring 0.5 inches on all sides, then how many cubes will he need?}
\end{mdframed}
\emph{Further Requirement}: Adding an additional requirement of transformation based on the output of the original function. \\
\begin{lstlisting}[language=Python, style=mystyle]
def flip_case_count(string: str) -> Tuple[str, int]:

    """
    For a given string, flip lowercase characters to uppercase and uppercase to lowercase. Additionally, return the number of case flips performed.

    >>> flip_case_count('Hello')
    ('hELLO', 5)
    """
\end{lstlisting}
\label{gs:subsequent-question}

\item \emph{Concurrent Question}: Adding an additional query or requirement that is independent from the original question. \\
\emph{Parallel Question}: Adding an additional \textbf{query} along with the original query based on the information given in the question, the added \textbf{query} should inquiry a value that is irrelevant of the original answer. Example:\\
\textbf{Changed from \ref{math-example:john}}:
\begin{mdframed}[backgroundcolor=pink!20] 
 John has 3 boxes. Each box is 5 inches by 6 inches by 4 inches. The walls are 1 inch thick. What is the total inner volume of all 3 boxes? \textbf{What is the total volume of the material used to build the boxes?} \\
\end{mdframed}
\emph{Parallel Requirement:} Adding an additional requirement in \textbf{Docstring} that does not rely on the output of the original question.\\
\emph{Changed from \ref{code-example:flip-case}}:
\begin{lstlisting}[language=Python, style=mystyle]
def flip_case_and_count(string: str) -> Tuple[str, int]:

    """For a given string, not only should you flip lowercase characters to uppercase and uppercase to lowercase. You should also output another Title case where only the first letter of each word is capitalized"""

    """>>> flip_case_and_count('Hello')
    ('hELLO', 'Hello')
    """
\end{lstlisting}
\label{gs:concurrent-question}

\item \emph{Change Question:} Change the current query or requirement to a different but similar one based on the existing information provided inside the question. \\
\emph{Change Query}: Change the \textbf{query} to ask for another value that requires more computation based on the information given in the question. \\
\textbf{Changed from \ref{math-example:john}}: 
\begin{mdframed}[backgroundcolor=pink!20] 
John has 3 boxes. Each box is 5 inches by 6 inches by 4 inches. The walls are 1 inch thick. What is the total \textbf{outer volume} of all 3 boxes? 
\end{mdframed}
\emph{Change Docstring}: Change the \textbf{Docstring} to another requirement based on the input given in the \textbf{question header}. \\
\textbf{Changed from \ref{code-example:derivative}}:
\begin{lstlisting}[language=Python, style=mystyle]
def calc_derivative(xs: list):

    """ xs represent coefficients of a polynomial.
    xs[0] * (exp (x))^0 + xs[1] * (exp(x))^1 + xs[2] * (exp(x))^2 + ....
    Return derivative of this polynomial in the same form.
    >>> derivative([3, 1, 2, 4, 5])
    [1, 4, 12, 20]
    >>> derivative([1, 2, 3])
    [2, 6]
    """
\end{lstlisting}
\label{gs:change-question}

\item \emph{Info Recombination}: Combine the fundamental concepts or frameworks from another question with the original question. \\
\emph{Info Recombination (Math)}: Graft \textbf{mathematical structure} from another question and combine with the original question. \\
\textbf{Changed from \ref{math-example:vicki}:} 
\begin{mdframed}[backgroundcolor=pink!20] 
Vicki and James are planning an event at their high school that combines a pop singing concert and dance events. The whole event will be 2 hours long. Vicki is allowing each musical group 2 minutes to get on stage, 6 minutes to perform, and then 2 minutes to exit the stage. James will also perform two solo dance routines, each lasting five minutes. Considering a 10-minute intermission during the show, how many musical groups can perform at the concert?
\end{mdframed}
\emph{Info Recombination (Code)}: Merge the requirement from another coding question with existing question.
\textbf{Changed from \ref{code-example:flip-case}}:
\begin{lstlisting}[language=Python, style=mystyle]
def flip_case_and_odd_sum(string: str) -> tuple:
    """
    Given a string, flip lowercase characters to uppercase and uppercase to lowercase.
    Also return the odd letters that are in even positions of the original string.
    string Index starts from 0, alphabet index start from 1. Aa is 1, Bb is 2..
    Examples:
    >>> flip_case_and_odd_sum('Hello')
    ('hELLO', 'o')
    """
\end{lstlisting}
\label{gs:info-recombination}

 \item \emph{Domain Knowledge}: Introduce a specific knowledge in math or code and merge it with the question. \\
 \emph{Theoretical Challenge}: Incorporate a specific theorem into the question so that perturbed question requires a new \textbf{toolbox} to solve. \\
\textbf{Changed from \ref{math-example:john}:} 
\begin{mdframed}[backgroundcolor=pink!20] 
John has an infinite number of boxes numbered as first, second, third, and so on. The first box is 5 inches by 6 inches by 7 inches in size. Starting from the second box each box is half the size of the previous box in each dimension. What is the total volume all the boxes combined? 
\end{mdframed}
\emph{Code Import:} The requirement requires to use a specific python library to solve the problem.
\textbf{Changed from \ref{code-example:flip-case}}:
Rewrite the function below to take in batch input parameters and use the multicore cpu for efficiency.
\label{gs:domain-knowledge}

\item \emph{Complex Reality}: Add an aspect of complexity in the real world scenario. \\
\emph{Value Probability}:Introduce concept of uncertainty to deterministic values and calculate the estimation. The perturbed question will require \textbf{toolbox} (knowledge) of probability. \\
\textbf{Changed from \ref{math-example:merchant}:} 
\begin{mdframed}[backgroundcolor=pink!20] 
A merchant wants to make a choice of purchase between 2 purchase plans: jewelry worth \$5,000 or electronic gadgets worth \$8,000. His financial advisor speculates that the jewelry market has a 70\% chance to go up 2.5\% and a 30\% chance to remain the same, while the electronic gadgets market will rise 1.2\% within the same month. If the merchant is looking to maximize profit at the end of this month by making a choice, how much estimated profit would this be? 
\end{mdframed}
\emph{Example Boundary}: Add boundary examples along with the existing \textbf{examples}. The boundary examples contains input that does not met requirement specified in the docstring.
\textbf{Changed from \ref{code-example:derivative}}:
Write a function to fulfill the requirement and all the examples inside the docstring 
\begin{lstlisting}[language=Python, style=mystyle]
def derivative(xs: list):

    """ xs represent coefficients of a polynomial.
    xs[0] + xs[1] * x + xs[2] * x^2 + ....
    Return derivative of this polynomial in the same form. The solution should pass all the test cases specified below

    # Regular case
    >>> derivative([3, 1, 2, 4, 5])
    [1, 4, 12, 20]
    # Smaller case
    >>> derivative([1, 2, 3])
    [2, 6]
    # Special case with empty list
    >>> derivative([])
    []
    # Boundary case, the shortest polynomial
    >>> derivative([1])
    [0]
    # Boundary case, all-zero polynomial
    >>> derivative([0.0, 0.0, 0.0])
    [0, 0]
    """
\end{lstlisting}
\label{gs:complex-reality}

\item \emph{General Solution}: Provide the solution in a more general scenario. \\
\emph{Code Implementation}: Develop a code function to solve the question in general. \\
\textbf{Changed from \ref{math-example:john}:} 
\begin{mdframed}[backgroundcolor=pink!20] 
\# Original Examples \#
Can you write a Python code to find out what is the total inner volume of all 3 boxes?\\
\end{mdframed}
\emph{Higher Order}: Write a higher order function that can solve the coding problem in general. \\
\textbf{Changed from \ref{code-example:gcd}}
\begin{lstlisting}[language=Python, style=mystyle]
def greatest_common_divisor(numbers: list[int]) -> int:
    """
    Calculates the greatest common divisor (GCD) of a list of integers.
    Returns the GCD as an integer.
    
    Examples:
    - For numbers = [20, 40, 60], the function returns 20.
    - For numbers = [35, 14], the function returns 7.
    """
\end{lstlisting}
\label{gs:general-solution}
\end{enumerate}

\item \textbf{Computation Adjustment}: While retaining the \textbf{Logical Structure}, this type aims to change one single reasoning step of the original question.

\begin{enumerate}[itemsep=0pt, leftmargin=*,  labelwidth=0pt, labelindent=0pt, parsep=0pt, label=\textbf{G\arabic*}.,resume]

\item \emph{Computation Demand}: Change the value to complex values that put a high demand on computation. \\
\emph{Value Big}: Significantly increasing the magnitude of values that pose a challenge for calculations. \\
\textbf{Changed from \ref{math-example:john}:} 
\begin{mdframed}[backgroundcolor=pink!20] 
John has 3000 boxes. Each box is 500 inches by 600 inches by 400 inches. The walls are 100 inches thick. 
What is the total inner volume of all the boxes? 
\end{mdframed}
\emph{Generalize Parameter}: Extend the current parameter into different python object types\\
\textbf{Changed from \ref{code-example:gcd}}:
\begin{lstlisting}[language=Python, style=mystyle]
def find_common_divisor(value1: Union[int, float, str], value2: Union[int, float, str]) -> float:
    """
    Takes two values (int, float, or float in string format) and finds the largest float that divides both into integers.
    Inputs can be a mix of types. Returns the divisor as a float.

    Examples:
    print(find_common_divisor("0.5", 1))  # 0.5
    print(find_common_divisor(0.25, "1.25"))  # 0.25
    """
\end{lstlisting}
\label{gs:computation-demand}

\item \emph{Change Value}: Change the content of the value to a different one. \\
\emph{Change Subject}: If there are multiple mentions in the question, Exchange \textbf{values} of names or references in the question. \\
\textbf{Changed from \ref{math-example:lily}}: 
\begin{mdframed}[backgroundcolor=pink!20] 
Together David, Bodhi, and Lily collected 43 insects. \textbf{David} found 7 more than \textbf{Bodhi}. \textbf{Bodhi} found half of what \textbf{Lily} found. How many insects did Lily find? 
\end{mdframed}
\emph{Parameter Content}: Change the format or meaning of the input parameter. \\
\textbf{Changed from \ref{code-example:derivative}}: \\
\begin{lstlisting}[language=Python, style=mystyle]
def derivative(polynomial: str):

    """ 'polynomial' is a string that stands for polynomial for form
    coefficients_0 + coefficients_1 * x + coefficients_2 * x^2 + ....
    This function will return the derivative of the aforementioned polynomial in the same format.

    >>> derivative('3 +1x + 2x^2 + 4x^3 + 5x^4')
    '1 + 4x + 12x^2 + 20x^3'
    >>> derivative('1 - 2x + 3x^2')
    '-2 + 6x'
    """
\end{lstlisting}
\label{gs:change-value}

\item \emph{Change Operation}: Change one operation regarding how the \textbf{Values} are processed.\\
\emph{Change Calculation}: Change no more than 3 words in original question so that the \textbf{toolbox} (mathematical operations) involved in the calculation are changed. \\
\textbf{Changed from \ref{math-example:john}}: 
\begin{mdframed}[backgroundcolor=pink!20] 
John has 3 boxes. The inner dimension of each box is 3 inches by 4 inches by 2 inches. The walls are 0.5 inches thick. What is the total outer volume of all 3 boxes? 
\end{mdframed}
\emph{Variable Type:} Change the python object type of the original parameter while keep its content the same, also specify the return variable to be in a specific type.  \\
\textbf{Changed from \ref{code-example:derivative}}:
\begin{lstlisting}[language=Python, style=mystyle]
def derivative(xs: list[str]) -> list[str]:

    """ xs represent coefficients of a polynomial.
    xs[0] + xs[1] * x + xs[2] * x^2 + ....
    Return derivative of this polynomial in the same form.
    """
\end{lstlisting}
\label{gs:change-operation}

\end{enumerate}
    
    \item \textbf{Formulation Adjustment}: This dimension tests the abstract reasoning ability of LLMs under the same logical structure of the original question. This dimension focuses on solving the general version of the original reasoning problem, rather than focusing on to get a standard solution. For math, We change the context to include one or more symbolic variables to replace its original \textbf{values}.

    \begin{enumerate}[itemsep=0pt, leftmargin=*,  labelwidth=0pt, labelindent=0pt, parsep=0pt, label=\textbf{G\arabic*}.,resume]
        
\item \emph{Symbolic Response}: Use logic to infer the final output after a sequence of steps. \\
\emph{Variable Response}: Replace one \textbf{value} inside the question with a variable and answer with the variable included.\\
\textbf{Changed from \ref{math-example:john}}: 
\begin{mdframed}[backgroundcolor=pink!20] 
John has X boxes. Each box is 5 inches by 6 inches by 4 inches. The walls are 1 inch thick. What is the total inner volume of all the boxes as a function of X? 
\end{mdframed}
{Code Execution}: Given Docstring requirement, and specific input parameter, find the output for the function without writing any code.
\textbf{Changed from \ref{code-example:flip-case}}:
Find the output of the following function description, if the input is:string = ``Hello World\!\&7''
\begin{lstlisting}[language=Python, style=mystyle]
def flip_case(string: str) -> str:
    """For a given string, flip lowercase characters to uppercase and uppercase to lowercase."""
\end{lstlisting}
\label{gs:symbolic-response}

\item \emph{Value Relationship}: Identify the relationship between input values or parameters if the output or the final answer is given. \\
\emph{Variable Relationship}: Replace a pair of \textbf{values} inside the question with variables. After answering the original question, the variable forms a relationship. Query that relationship.\\
\textbf{Changed from \ref{math-example:john}}: 
\begin{mdframed}[backgroundcolor=pink!20] 
John has X boxes. Each box is Y inches by 6 inches by 4 inches. The walls are 1 inch thick. If the total inner volume of all the boxes is 72 cubic inches, then find the equation that relates X and Y? 
\end{mdframed}
\emph{Parameter Relationship:} Given the output of the function, categorize the possible groups of inputs parameters into the question.
\textbf{Changed from \ref{code-example:gcd}:}
If the below program output integer 7. What is the relationship between a and b
\begin{lstlisting}[language=Python, style=mystyle]
def function(a: int, b: int) -> int:
    while b:
        a, b = b, a % b
    return a
\end{lstlisting}
\label{gs:values-relationship}

\item \emph{Variable Group}: Change a group of several input values or parameters to variables.\\
\emph{Variable Scaling}: After answering the question, change the \textbf{query} to: if certain factual numbers in the question is scaled up by x, how will the final answer change?\\
\textbf{Changed from \ref{math-example:john}}: 
\begin{mdframed}[backgroundcolor=pink!20] 
John has 3 boxes. Each box is 5 inches by 6 inches by 4 inches. The walls are 1 inch thick. Now, the number of boxes, the box outer dimensions, and the wall thickness are all scaled up by a factor of X. What is the total inner volume of all the boxes as a function of X? 
\end{mdframed}
\emph{Variable Substitution}: Change one or more variables inside the docstring to input parameters. 
\textbf{Changed from \ref{code-example:flip-case}}:
\begin{lstlisting}[language=Python, style=mystyle]
def flip_case(string: str, specific_value: str) -> str:

    """"""For a given string and specific value, flip the specific value from lowercase to uppercase or uppercase to lowercase.  The function will only flip the case of the specific value in the string.
    >>> flip_case('Hello', 'h')
    'hello'
    """
\end{lstlisting}
\label{gs:variable-group}

\item  \emph{Backward Reasoning}: Reverse the reasoning process, reason from how to reach input from output. \\
\emph{Variable Adaptation}: If the answer to the question add or subtract by a certain amount x, pick one \textbf{value} inside the \textbf{Information} and ask how it should change if other \textbf{values} are kept the same.\\
\textbf{Changed from \ref{math-example:john}}:
\begin{mdframed}[backgroundcolor=pink!20]  
John has 3 boxes. Each box is 5 inches by 6 inches by 4 inches. The walls are 1 inch thick. If the total inner volume of all 3 boxes increases by a certain variable X, how should the thickness of the walls adjust correspondingly if the number of boxes and the external dimensions of each box stay the same? Write the answer as a function of X. 
\end{mdframed}
\emph{Reverse Engineering}: Change the \textbf{Docstring, Function Header, and Examples} to find the function that can reverse engineer the original function. Specifically, mapping the output back to its original inupt.
\textbf{Changed from \ref{code-example:flip-case}}:
Create a function that reverses the following function's process, effectively transforming its output back into the original input
\begin{lstlisting}[language=Python, style=mystyle]
def function(string: str) -> str:
    return string.swapcase()
\end{lstlisting}
\label{gs:backward-reasoning}

\item \emph{Counterfactual}: What would the outcome be if X had happened instead of Y, given the same initial conditions and context. \\
\emph{WhatIf Question}: First mask some number of \textbf{values} inside the question and answer the original question. What if we change one value inside the question, how will the final answer change? (The final answer should not have variables included as the masked value could be solved given the final answer.)\\
\textbf{Changed from \ref{math-example:john}:} 
\begin{mdframed}[backgroundcolor=pink!20] 
John has 3 boxes. Each box is 5 inches in width by 6 inches in length and a few inches in height. The walls are 1 inch thick. The total inner volume of all the boxes combined is 72 cubic inches. Now, if the thickness of the walls is half of its original thickness, then what will be the total inner volume? 
\end{mdframed}
\emph{WhatIf Code}: WhatIf the \textbf{code structure} or \textbf{input value} is changed, and some condition is masked.
\textbf{Changed from \ref{code-example:flip-case}}:
Find the output of the `changed\_function`, if the input is the same.
\begin{lstlisting}[language=Python, style=mystyle]
We know that if we input masked_input to the `original_function`, the output is following:
>>> original_function(masked_input)
'hELLO'

Here is the `original_function`:
def original_function(string: str) -> str:
    return string.swapcase()

Here is the `changed_function`:
def changed_function(string: str) -> str:
    return string.swapcase()[::-1]

What will be the output for `changed_function(masked_input)`"
\end{lstlisting}
\label{gs:what-if}
        
\item \emph{Solve Value}: Mask one variable's \textbf{value} inside question, given answer, infer the masked value.\\
\emph{Solve X}: Replace one value inside the question with X and solve for X.\\
\textbf{Changed from \ref{math-example:john}:} 
\begin{mdframed}[backgroundcolor=pink!20] 
John has X boxes. Each box is 5 inches by 6 inches by 4 inches. The walls are 1 inch thick. If the total inner volume of all 3 boxes is 72 cubic inches what is the value for X?
\end{mdframed}
\emph{Solve Input:} Determine the input value of the function, based on the known output value. \\
\textbf{Changed from \ref{code-example:flip-case}}
What is input to the following function, if the output is: "hELLO 9"
\begin{lstlisting}[language=Python, style=mystyle]
def function(string: str) -> str:
    return string.swapcase()
\end{lstlisting}
\label{gs:solve-value}

\item \emph{Identify Range}: Find what are possible constraint on the values.\\
\emph{Variable Range}: Replace one \textbf{value} with variable, and change the \textbf{query} to find the possible range of values based on the question.\\
\textbf{Changed from \ref{math-example:john}}: 
\begin{mdframed}[backgroundcolor=pink!20] 
John has 3 boxes. Each box is X inches by 6 inches by 4 inches. The walls are 1 inch thick. Suppose we want to find out the total inner volume of all the boxes. What are the possible ranges of values of variable X based on the given information? 
\end{mdframed}
\emph{Parameter Range}: Identify what are the constraint on the input parameter, or what is the range of output parameter if input parameter is contraint to take certain value. \\
\textbf{Changed from \ref{code-example:derivative}}:
If all the item inside the input list is smaller than 1, what are the constraints on the output from this function below?
\begin{lstlisting}[language=Python, style=mystyle]
def function(xs: list):
    return [(i * x) for i, x in enumerate(xs)][1:]
\end{lstlisting}
\label{gs:identify-range}
\end{enumerate}
\end{enumerate}

\subsection{Concept Analysis}
\label{sec:open-question}
This perturbation type encompasses questions that concentrate on the model's capabilities beyond mere problem-solving accuracy. The responses to these questions should be in natural language format. Instead of just assessing whether the model can correctly predict answers to new questions, we aim to examine the depth of knowledge the models possess and understanding of important concepts and rationales in the process of solving the original questions. Essentially, we are asking: \emph{Does the model predict correctly because it truly understands the question?} To test this, we observe how the model behaves in different or unusual scenarios that are not typically presented in standard questions.

\begin{enumerate}[label=(\roman*), leftmargin=*, wide, labelwidth=0pt, labelindent=0pt]
\item \textbf{Question Understanding}: This dimension examines how model decompose, interpret and analyze the information inside the question.
\begin{enumerate}[itemsep=0pt, leftmargin=*,  labelwidth=0pt, labelindent=0pt, parsep=0pt, label=\textbf{G\arabic*}.,resume]
\setcounter{enumii}{22}

\item \emph{Inherent Premise}: Identify the underlying premise of the question. \\
\emph{Identify Assumption}: Identify one hidden commonsense assumption in the question that requires the answer to be answerable. \\
\textbf{Changed from \ref{math-example:kylar}:} 
\begin{mdframed}[backgroundcolor=pink!20] 
You do not need to solve the question below, just identify one important hidden assumption that is required for the question to be answerable.
Kylar went to the store to buy glasses for his new apartment. One glass costs \$5, but every second glass costs only 60\% of the price. Kylar wants to buy 16 glasses. How much does he need to pay for them? 
\end{mdframed}
\emph{Test Case}: List different boundary test cases that is valid for the input of the question. 
\textbf{Changed from \ref{code-example:flip-case}}:
Provide input parameters for the test cases of the specified coding problem. These parameters should encompass boundary conditions within the scope defined by the function's requirements specification, and avoid scenarios that fall outside of these requirements.
\begin{lstlisting}[language=Python, style=mystyle]
def flip_case(string: str) -> str:
    """For a given string, flip lowercase characters to uppercase and uppercase to lowercase.
    """
\end{lstlisting}
\label{gs:inherent-premise}

\item \emph{Complete Missing}: Fulfill the missing information in the question by analyze how the information is structured and presented inside the question. \\
\emph{Info Sufficiency}: Mask or delete an important piece of \textbf{information} and ask what additional information is needed to make the question answerable.\\
\textbf{Changed from \ref{math-example:john}:} 
\begin{mdframed}[backgroundcolor=pink!20] 
John owns 3 boxes, each measuring 5 inches by 6 inches by 4 inches. Each box also had inner walls with certain non-zero thicknesses. Suppose you want to find out the total inner volume of all the boxes. What information is missing to calculate that? 
\end{mdframed}
\emph{Incomplete Answer}: Given the question, mask partial answer of the original, the model need to infer the missing lines based on the context.
\textbf{Changed from \ref{code-example:flip-case}}:
Complete the function below by predicting what is inside the masked code paragraph
\begin{lstlisting}[language=Python, style=mystyle]
def flip_case(string: str) -> str:
    """For a given string, flip lowercase characters to uppercase and uppercase to lowercase.
    >>> flip_case('Hello')
    'hELLO'
    """
    [masked code paragraph]
        if char.isupper():
            result += char.lower()
        else:
            result += char.upper()
    return result
\end{lstlisting}
\label{gs:complete-missing}

\item \emph{Question Formulation}: Formulate the question based on its answer.\\
\emph{Question Formulation - (Math)}: Formulate a \textbf{question} to the chain of thought \textbf{gold answer}.\\ 
\textbf{Changed from \ref{math-example:merchant}}:
\begin{mdframed}[backgroundcolor=pink!20] 
Formulate a math application \textbf{question} that requires the following \textbf{mathematical structure} (calculations):
5000*(2.5/100) = \$125
8000*(1.2/100) = \$96
\$125 > \$96
\$125
Math Question: Ask potential structures of math application.
\end{mdframed}
\emph{Question Formulation - (Code)}: Formulate a concise coding requirement by looking at the function code.\\
\textbf{Changed from \ref{code-example:gcd}}:
Write a concise code description for the following code of its functionality no more than 1 sentence.
\begin{lstlisting}[language=Python, style=mystyle]
def function(a,b):
    while b:
        a, b = b, a % b
    return a
\end{lstlisting}
\label{gs:question-formulation}

\item \emph{Add Misinformation}: Add a piece of distracting information that can mislead the answer.\\
\emph{Introduce Distraction}: Add a Potentially Distracting \textbf{information} that will not affect the answer to the question.\\
\textbf{Changed from \ref{math-example:merchant}: }
\begin{mdframed}[backgroundcolor=pink!20] 
A merchant is considering a decision between the following purchase plans: jewelry with a value of \$5,000, a trip to Europe costing \$7,000, or electronic gadgets worth \$8,000. His financial advisor predicts that the jewelry market will increase by 2.5\%, the travel market will stay relatively stable with little to no change, and the electronic gadgets market will rise by 1.2\% within the same month. He recently also came into an inheritance of \$20,000 that he doesn't need to use right away. If the merchant's goal is to maximize profit at the end of this month by making a purchase choice, how much profit would this be? 
\end{mdframed}
\emph{Introduce Bias:} Change the python header to describe another function requirement, and change all the examples demonstrations bias towards a specific output
\textbf{Changed from \ref{code-example:flip-case}}
\begin{lstlisting}[language=Python, style=mystyle]
def uppercase(string: str) -> str:
    """For a given string, flip lowercase characters to uppercase and uppercase to lowercase.
    >>> flip_case('hello')
    'HELLO'
    """
\end{lstlisting}
\label{gs:add-misinformation}
 \end{enumerate}

\item \textbf{Solution Understanding}: This dimension focuses on the problem-solving process to get to the final answer and how to optimize it.  
\begin{enumerate}[itemsep=0pt, leftmargin=*,  labelwidth=0pt, labelindent=0pt, parsep=0pt, label=\textbf{G\arabic*}.,resume]

\item \emph{Optimize Solution}: Assess whether the current state is optimal or if improvements are necessary.\\
\emph{Info Necessity}: Check If there is redundant \textbf{information} given in the question, if yes, identify the redundant information.\\
\textbf{Changed from \ref{math-example:john}:} 
\begin{mdframed}[backgroundcolor=pink!20] 
John has 3 boxes. Each box is 5 inches by 6 inches by 4 inches. The walls are 1 inch thick. Suppose we want to find out the total inner volume of all 3 boxes. To solve this math question, is there a way to determine the total inner volume of all 3 boxes without calculating the inner volume of one box?
\end{mdframed}
\emph{Reduce Complexity}: Assess whether the complexity of the current code be further reduced. \\
\textbf{Changed from \ref{code-example:derivative}}:
Optimize the code below to more efficiently achive the same requirement specified in the docstring
\begin{lstlisting}[language=Python, style=mystyle]
def derivative_polynomial(coefficients, derivative=None, index=0):
    """
    This function calculates the derivative of a polynomial using recursion.
    coefficients: List of coefficients of the polynomial.
    derivative: List to store the coefficients of the derivative. Initially None.
    index: Current index in the coefficients list.
    
    The base case of the recursion is when the index is equal to the length of the coefficients list.
    """
    if index > 0:
        derivative_coefficient = index * coefficients[index]
        derivative.append(derivative_coefficient)
    return derivative_polynomial(coefficients, derivative, index + 1)
\end{lstlisting}
\label{gs:optimize-solution}

\item \emph{Step Functionality}: Whether there are alternative answers that follow the constraint. \\
\emph{Step Necessity}: Whether there are any alternative solutions \textbf{reasoning steps} without calculating an specific intermediate value. \\
\textbf{Changed from \ref{math-example:john}:} 
\begin{mdframed}[backgroundcolor=pink!20] 
John has 3 boxes. Each box is 5 inches by 6 inches by 4 inches. The walls are 1 inch thick. Suppose we want to find out the total inner volume of all 3 boxes. To solve this math question, is there a way to determine the total inner volume of all 3 boxes without calculating the inner volume of one box? 
\end{mdframed}
\emph{Step Necessity:} Provide one line of code inside the Python function, and explain the functionality of that line of code in the context of the whole solution.\\
\textbf{Changed from \ref{code-example:gcd}}:
Explain what is the the line below the comment functionality?
\begin{lstlisting}[language=Python, style=mystyle]
def greatest_common_divisor(a: int, b: int) -> int:

    """ Return a greatest common divisor of two integers a and b
    >>> greatest_common_divisor(3, 5)
    1
    >>> greatest_common_divisor(25, 15)
    5
    """
    while b:
        a, b = b, a % b
    # What is the functionality of `abs()`
    return abs(a)
\end{lstlisting}
\label{gs:step-functionality}

% \item \emph{Alternative Question}: Generate the question in a different scenario without changing the structure of the question.
% \emph{Alternative Question (Math)}: Rephrase the Question in another application setting, but keep the \textbf{mathematical structure} intact. This question focuses on finding a different application scenario that shares the same mathematical structure. \\
% \textbf{Original}: John has 3 boxes. Each box is 5 inches by 6 inches by 4 inches. The walls are 1 inch thick. What is the inner volume of all 3 boxes?\\
% \textbf{Changed}: Rephrase the question in a completely different scenario without changing the mathematical core structure of the question. You do not need to answer the question.
%  John has 3 boxes. Each box is 5 inches by 6 inches by 4 inches. The walls are 1 inch thick. What is the inner volume of all 3 boxes? \\
%  \emph{Alternative Question (Code)}:  
% \label{gs:alternative-question}

\item \emph{Theoretical Basis}: Identify the theory or principles in solving the question in general. \\
\emph{Theoretical Basis (Math)}: Identify the underlying arithmetic or algebraic rules (\textbf{toolbox}) that govern the solution to the question. \\
\textbf{Changed from \ref{math-example:john}:} 
\begin{mdframed}[backgroundcolor=pink!20] 
John has 3 boxes. Each box is 5 inches by 6 inches by 4 inches. The walls are 1 inch thick. Assume you want to find out the total inner volume of all 3 boxes. Can you identify one underlying mathematical theory which is required to do that? 
\end{mdframed}
\emph{Theoretical Basis (Code)}: Request explanation on essential python concepts required to solve the question, for example, related to python objects and programming skills. \\
\textbf{Changed from \ref{code-example:flip-case}}:
Please describe to me in simple terms, assuming I have no knowledge of programming. Your task isn't to solve the coding problem itself, but rather to identify the programming concepts in Python that would be necessary to address the problem presented below.
\begin{lstlisting}[language=Python, style=mystyle]
def flip_case(string: str) -> str:
    """For a given string, flip lowercase characters to uppercase and uppercase to lowercase.
    >>> flip_case('Hello')
    'hELLO'
    """
\end{lstlisting}
\label{gs:theoretical-basis}

% \item \emph{Identical Question}: Check if the two questions are identical in their \textbf{mathematical structure.}\\
% \textbf{Original}: A merchant wants to make a choice of purchase between 2 purchase plans: jewelry worth \$5,000 or electronic gadgets worth \$8,000. His financial advisor speculates that the jewelry market will go up 2.5\% while the electronic gadgets market will rise 1.2\% within the same month. If the merchant is looking to maximize profit at the end of this month by making a choice, how much profit would this be? \\
% \textbf{Changed}: Question 1: A merchant wants to make a choice of purchase between 2 purchase plans: jewelry worth \$5,000 or electronic gadgets worth \$8,000. His financial advisor speculates that the jewelry market will go up 2.5\% while the electronic gadgets market will rise 1.2\% within the same month. If the merchant is looking to maximize profit at the end of this month by making a choice, how much profit would this be? Question 2: An investor is unsure of which investment to make: gold valued at \$10,000 or stocks valued at \$15,000. His financial consultant predicts that the gold market will inflate by 3.5\% while the stock market will increase by 2.2\% over the next quarter. If the investor wants to achieve the highest return on his investment at the end of this quarter, how much would his initial investment be? Does Question 1 and Question 2 require identical steps to answer? \\
% \label{gs:identical-question}

\item \emph{Cost Analysis}: Analyze the computational cost regarding the solution. \\
\emph{Solution Efficiency}: Compare two solution plans on solving the question and evaluate which one uses less computation. \\
\textbf{Changed from \ref{math-example:john}}:
\begin{mdframed}[backgroundcolor=pink!20] 
Evaluate which solution plan is more efficient in solving the question?

Plan 1: Calculate the volume of the outer dimensions for one box, calculate the volume of the material used for the walls for one box, subtract the latter from the former to find the inner volume of one box, and then multiply this by 3 for all boxes.

Plan 2: Calculate the inner dimensions of a single box by subtracting twice the thickness of the walls from each outer dimension, then find the volume of this inner space and multiply by 3 for all boxes.
\end{mdframed}
\emph{Code Complexity}: Analyze the time complexity and space complexity of the provided code solution. \\
\textbf{Changed from \ref{code-example:flip-case}}
Analyze the time and space complexity regarding to input parameter string of the following function.
\begin{lstlisting}[language=Python, style=mystyle]
def flip_case(string: str) -> str:
    """For a given string, flip lowercase characters to uppercase and uppercase to lowercase.
    >>> flip_case('Hello')
    'hELLO'
    """
\end{lstlisting}
\label{gs:cost-analysis}

\end{enumerate}
% \end{enumerate}
% Through these dimensions, we aim to understand and measure the language model's proficiency in not only solving problems but also in following specific directions and adapting its responses according to varied requirements.

% \subsubsection{Robustness}
% \label{sec:robustness}
% In the Robustness Domain, our focus is on evaluating how well Large Language Models (LLMs) maintain accuracy when faced with intentional modifications to the original questions. These modifications are designed to test the LLMs' susceptibility to errors. Here's a clearer description of the dimensions within this domain:
% \begin{enumerate}[label=(\roman*), leftmargin=*, wide, labelwidth=0pt, labelindent=0pt]
%     \item \textbf{Change Format}: This dimension is inspired by metamath \hl{cite metamath}. It involves rephrasing or restructuring the same question in different ways. The objective is to verify whether the LLM can still provide correct answers even when the format or presentation of the question changes. This tests the model's ability to understand the underlying concept irrespective of how it's presented.
%     \begin{enumerate}[itemsep=0pt, leftmargin=*,  labelwidth=0pt, labelindent=0pt, parsep=0pt, label=\textbf{G\arabic*}.,resume]
%     \setcounter{enumii}{34}

%     \end{enumerate}
\item \textbf{Critical Thinking}: In this dimension, deliberate errors are introduced into the question or in a provided example answer. The purpose is to see if the LLM can identify and rectify these errors. This tests the LLM's error detection capabilities, which is crucial for reliability in practical applications.
\begin{enumerate}[itemsep=0pt, leftmargin=*,  labelwidth=0pt, labelindent=0pt, parsep=0pt, label=\textbf{G\arabic*}.,resume]

\item \emph{Seek Clarification}: The question requires to be clarified first before answering.\\ 
\emph{Introduce Ambiguity}: Introduce Ambiguity to the question implicitly by changing the original \textbf{information}, so that the question cannot be solved without clarification. \\
\textbf{Changed from \ref{math-example:john}:} 
\begin{mdframed}[backgroundcolor=pink!20] 
John has three 5x6x4 inch boxes. A particular side of each box have 1 inch thick walls. What does the total inner capacity of these boxes amount to? 
\end{mdframed}
\emph{Example Requirement}: Remove the coding requirement in the docstring, instead only provide examples as a coding requirement. The provided examples will define and demonstrate the expected behavior in various scenarios.
\textbf{Changed from \ref{code-example:flip-case}}:
Begin by analyzing the function's behavior specified in the docstring to understand its pattern, and then proceed to code the function accordingly.
\begin{lstlisting}[language=Python, style=mystyle]
def flip_case(string: str) -> str:
    """
    function('Hello') == 'hELLO'
    function('Python 3.8') == 'pYTHON 3.8'
    function('123abcXYZ') == '123ABCxyz'
    function('MixedCASE123') == 'mIXEDcase123'
    function('ALLUPPERCASE') == 'alluppercase'
    """
\end{lstlisting}
\label{gs:seek-clarification}

\item \emph{Conditional Analysis}: Based on different possible situations of the question, the answer should separately presented. \\
\emph{Discuss Separately}: Introduce new \textbf{information} containing variables or conditions that require the answer to be discussed separately based on conditions or variables. \\
\textbf{Changed from \ref{math-example:merchant}}: 
\begin{mdframed}[backgroundcolor=pink!20] 
A merchant wants to make a choice of purchase between 2 purchase plans: jewelry worth \$5,000 or electronic gadgets worth \$8,000. His financial advisor speculates that the jewelry market will go up x\% while the electronic gadgets market will rise 1.2\% within the same month. If the merchant is looking to maximize profit at the end of this month by making a choice, how much profit would this be?  
\end{mdframed}
\emph{Incomplete requirement:} Left some condition unspecified in the docstring.
\textbf{Changed from \ref{code-example:flip-case}}:
\begin{lstlisting}[language=Python, style=mystyle]
def flip_case(ch: str) -> str:

    """For a given string, all the letters inside the string should be changed. flip lowercase characters to uppercase.
    >>> flip_case('h')
    'H'
    """
\end{lstlisting}
\label{gs:conditional-analysis}

\item \emph{Conflicting Information}: Introduce a new piece of information that is conflicting with existing information. This will make the question unanswerable, so the if the LLM can spot the error without mentioning.\\
\emph{Introduce Contradiction}: Add a piece of contradicting \textbf{information} to the question and check if LLM can spot the problem.\\
\textbf{Changed from \ref{math-example:john}: }
\begin{mdframed}[backgroundcolor=pink!20] 
John has 3 boxes. Each box is 5 inches by 6 inches by 4 inches. The walls are 1 inch thick. Each box is also 8 inches in width. What is the total inner volume of all 3 boxes? 
\end{mdframed}
\emph{Wrong Example}: Include an example that is conflicting with the requirement specified in the docstring. 
\textbf{Changed from \ref{code-example:flip-case}}:
\begin{lstlisting}[language=Python, style=mystyle]
def flip_case(string: str) -> str:
    """"""For a given string, flip lowercase characters to uppercase and uppercase to lowercase.
    >>> flip_case('Hello')
    'hello'
    """
\end{lstlisting}
\label{gs:conflicting-information}

\item \emph{Surface Error}: Introduce an obvious error that can be spotted without reasoning.\\
\emph{Value Uncommon}: Change the \textbf{values} so that it seems weird or unusual by commonsense knowledge standards.\\
\textbf{Changed from \ref{math-example:john}:}
\begin{mdframed}[backgroundcolor=pink!20] 
Can you spot anything unusual in the following question? 
John has 3 boxes. Each box measures 50000 miles by 60000 miles by 40000 miles. The walls of the boxes are 100 miles thick. What is the total inner volume of all 3 boxes? 
\end{mdframed}
\emph{Runtime Error}: Introduce a piece of error that will cause a runtime error or syntax error in python.
\textbf{Changed from \ref{code-example:flip-case}}:
Debug the error in the following code
\begin{lstlisting}[language=Python, style=mystyle]
def flip_case(string, str) -> str:
    """For a given string, flip lowercase characters to uppercase and uppercase to lowercase.
    >>> flip_case('Hello')
    'hELLO'
    """
    return string.swapcase()
\end{lstlisting}
\label{gs:surface-error}

\item \emph{Hidden Error}: Introduce a hidden error that need logical reasoning to spot.\\
\emph{Value Error}: Change the \textbf{values} so that the question does not make sense. \\
\textbf{Changed from \ref{math-example:vicki}:} 
\begin{mdframed}[backgroundcolor=pink!20] 
Vicki is planning a pop concert at her high school. The show will be 2 minutes. She is allowing each group 2 hours to get on stage, 6 hours to perform, and then 2 hours to exit the stage. If she allows a 10-hour intermission, how many groups can perform in the concert?
\end{mdframed}
\emph{Logical Error}: Introduce the change in the code that will cause a Logical Error in Python.
\textbf{Changed from \ref{code-example:flip-case}}
\begin{lstlisting}[language=Python, style=mystyle]
def flip_case(string: str) -> str:
    """For a given string, flip lowercase characters to uppercase and uppercase to lowercase.
    >>> flip_case('Hello')
    'hELLO'
    """
    string = list(string.swapcase())
    return string
\end{lstlisting}
\label{gs:hidden-error}

\end{enumerate}

\end{enumerate}

\subsection{Question Format}
\label{sec:format-change}
This dimension is inspired by metamath~\citep{yu2023metamath}. It involves changing the \textbf{question representation} by modifying the question encoding or specify the representation of the answer in different ways while keeping the underlying logical structure and conceptual framework of the original question intact. The objective is to verify whether the LLM can still provide correct answers even when the format or presentation of the question changes. This tests the model's ability to reason irrespective of how it's presented. It also tests models' instruction following ability where the \textbf{answer representation} must follow a certain format.
\begin{enumerate}[label=(\roman*), leftmargin=*, wide, labelwidth=0pt, labelindent=0pt]
\item  \textbf{Format Change}:
\begin{enumerate}[itemsep=0pt, leftmargin=*,  labelwidth=0pt, labelindent=0pt, parsep=0pt, label=\textbf{G\arabic*}.,resume]
\setcounter{enumii}{35}

\item \emph{Setting Rephrase}: Rephrase the question in another setting.\\
\emph{Change Setting}: Rephrase by changing the application setting and values inside the information, while keeping the core mathematical structure intact.\\
\textbf{Changed from \ref{math-example:john}}: 
\begin{mdframed}[backgroundcolor=pink!20] 
Maria has 4 cuboids. Each cuboid is 7 feet by 9 feet by 6 feet. The walls are 2 feet thick. What is the total volume of all the cuboids? 
\end{mdframed}
\emph{Realworld Usecase}: Frame the requirement in docstring into a problem that will happen in a realworld scenario.
\textbf{Changed from \ref{code-example:flip-case}}:
\begin{lstlisting}[language=Python, style=mystyle]
def switch_text_case(text: str) -> str:
    """
    Imagine you're working on a document and you've mistaken the case in the text you write. You wrote all the lower case letters in uppercase and vice versa, suppose you want to correct all of them using python. 
    """
\end{lstlisting}
\label{gs:setting-rephrase}

\item \emph{Change Sequence}: Change the order of the information and names of the variables that is originally presented in the question. \\
\emph{Change Sequence}: Change the sequence of information given in the question without affecting the solvability of the question. \\
\textbf{Changed from \ref{math-example:john}}:
\begin{mdframed}[backgroundcolor=pink!20] 
The walls of John's boxes are 1 inch thick. Each of these boxes measures 5 inches by 6 inches by 4 inches. John has 3 boxes. What is the total inner volume of all 3 boxes?
\end{mdframed}
\emph{Parameter Sequence}: Change the sequence of the input parameter and change the input parameter names.\\
\textbf{Changed from \ref{code-example:gcd}}
\begin{lstlisting}[language=Python, style=mystyle]
def munchee_bunchee(xray: int, yoyo: int) -> int:

    """ Return a common divisor that is the largest of two integers xray and yoyo
    >>> munchee_bunchee(3, 5)
    1
    >>> munchee_bunchee(25, 15)
    5
    """
\end{lstlisting}
\label{gs:change-sequence}

\item \emph{Close Format}: Rewrite the sentence as a closed-format question that evaluates the correctness of possible answers. \\
\emph{True False}: Evaluate a potentially misleading answer and check the correctness of the answer. \\
\textbf{Changed from \ref{math-example:merchant}}: 
\begin{mdframed}[backgroundcolor=pink!20] 
A merchant wants to make a choice of purchase between 2 purchase plans: jewelry worth \$5,000 or electronic gadgets worth \$8,000. His financial advisor speculates that the jewelry market will go up 2.5\% while the electronic gadgets market will rise 1.2\% within the same month. If the merchant is looking to maximize profit at the end of this month by making a choice, how much profit would this be? Evaluate the correctness of this answer with respect to the above question: \$96. 
\end{mdframed}
\emph{True False}: Check if a given code answer can solve the requirement in docstring.
\textbf{Changed from \ref{code-example:gcd}}:
Evaluate whether the solution below is the correct solution for the coding question, True or False?
\begin{lstlisting}[language=Python, style=mystyle]
Function:

def greatest_common_divisor(a: int, b: int) -> int:

    """ Return a greatest common divisor of two integers a and b
    >>> greatest_common_divisor(3, 5)
    1
    >>> greatest_common_divisor(25, 15)
    5
    """


Solution:

    while a:
        a, b = a % b, a
    return b
\end{lstlisting}
\label{gs:close-format}

\item \emph{Data Restructuring}: Change the layout, organization of the data presented in the question. \\
\emph{Value Structuring}: Arrange the variables inside the question in a tabular format.\\
\textbf{Changed from \ref{math-example:john}:} 
\begin{center}
\begin{verbatim}
| Variable | Value |
|----------|-------|
| a        | 3     |
| b        | 5     |
| c        | 6     |
| d        | 4     |
| e        | 1     |
\end{verbatim}
\end{center}
John has `a' boxes. Each box is `b' inches by `c' inches by `d' inches in dimensions. The walls are `e' inch thick. What is the total inner volume of all the `a' boxes? \\
\emph{Complex Docstring}: Elaborate the documentation string by exhaustively detailing more conditional pathway within the code. \\
\textbf{Changed from \ref{code-example:flip-case}}:
\begin{lstlisting}[language=Python, style=mystyle]
def function(string: str = None) -> str: 
    """
    For any specified sequence of alphabetical characters, interspersed with spaces, numerical digits, and various symbols, implement a sophisticated transformation algorithm designed to selectively convert  each alphabetical character from its current case representation, either lowercase or uppercase, to its diametrically opposite case representation. This algorithm ensures that every character initially presented in lowercase is meticulously transmuted to uppercase, and conversely, every character originally in uppercase is converted to lowercase, while meticulously preserving the integrity and original positioning of spaces, numerical digits, and any other non-alphabetical symbols, leaving these elements unaltered within the sequence.
    >>> function('Hello')
    'hELLO'
    """
\end{lstlisting}
\label{gs:data-structuring}
\end{enumerate}

\item  \textbf{Format Comparison}:
\begin{enumerate}[itemsep=0pt, leftmargin=*,  labelwidth=0pt, labelindent=0pt, parsep=0pt, label=\textbf{G\arabic*}.,resume]
\setcounter{enumii}{39}
\item \emph{Identical Problem}: Check if the two question or code are identical in describing or solving the same problem. \\
\emph{Identical Question:} If two questions requires exactly the same framework or thinking procedure to solve. \\
\textbf{Changed from \ref{math-example:merchant}}: 
\begin{mdframed}[backgroundcolor=pink!20] 
Question 1: A merchant wants to make a choice of purchase between 2 purchase plans: jewelry worth \$5,000 or electronic gadgets worth \$8,000. His financial advisor speculates that the jewelry market will go up 2.5\% while the electronic gadgets market will rise 1.2\% within the same month. If the merchant is looking to maximize profit at the end of this month by making a choice, how much profit would this be? 
Question 2: An investor is unsure of which investment to make: gold valued at \$10,000 or stocks valued at \$15,000. His financial consultant predicts that the gold market will inflate by 3.5\% while the stock market will increase by 2.2\% over the next quarter. If the investor wants to achieve the highest return on his investment at the end of this quarter, how much would his initial investment be? Does Question 1 and Question 2 require identical steps to answer?
\end{mdframed}
\emph{Identical Code:} Are the two solutions to the question identical in terms of their functionality?\\
\textbf{Changed from \ref{code-example:derivative}}
Is function\_1 and function\_2 identical in terms of its functionality?
\begin{lstlisting}[language=Python, style=mystyle]
Code 1:
def function(xs: list):
    return [(i * x) for i, x in enumerate(xs)][1:]
Code 2:
def function(xs: list):
    derivative = [i * xs[i] for i in range(1, len(xs))]
\end{lstlisting}
\label{gs:identical-problem}
\end{enumerate}
\end{enumerate}
\subsection{Answer Format}
\begin{enumerate}[label=(\roman*), leftmargin=*, wide, labelwidth=0pt, labelindent=0pt]
\item \textbf{Answer Constraint}: This dimension add a constraint on the solution so that it should conduct reasoning under the constraint
\begin{enumerate}[itemsep=0pt, leftmargin=*,  labelwidth=0pt, labelindent=0pt, parsep=0pt, label=\textbf{G\arabic*}.,resume]
\setcounter{enumii}{40}
\item \emph{Reasoning Format}: The format for the final answer should be converted in a certain way.\\ 
\emph{Binary Coded}: Answer the final question in base-n.\\
\textbf{Changed from \ref{math-example:john}:}
\begin{mdframed}[backgroundcolor=pink!20] 
Answer the following question with only base-2 coded values. 
Question: John has 3 boxes. Each box is 5 inches by 6 inches by 4 inches. The walls are 1 inch thick. What is the total inner volume of all 3 boxes?
\end{mdframed}
\emph{No Keyword}: The solution should not use a specific python keyword. For example, ``for'' or ``while''.
\textbf{Changed from \ref{code-example:gcd}}:
Answer the coding function below without using python keywords: "while", "for" in the solution
\begin{lstlisting}[language=Python, style=mystyle]
def greatest_common_divisor(a: int, b: int) -> int:

    """ Return a greatest common divisor of two integers a and b
    >>> greatest_common_divisor(3, 5)
    1
    >>> greatest_common_divisor(25, 15)
    5
    """
\end{lstlisting}
\label{gs:reasoning-format}

\item \emph{Reasoning Style}: The reasoning steps should be performed in a certain style. \\
\emph{X Language (Math)}: Give the answer in certain language from {Spanish, Chinese, Bengali, English, French} \\
\textbf{Changed from \ref{math-example:john}:} Answer the following question with only Chinese language, because I do not understand English. \\
\begin{mdframed}[backgroundcolor=pink!20] 
Question: John has 3 boxes. Each box is 5 inches by 6 inches by 4 inches. The walls are 1 inch thick. What is the total inner volume of all 3 boxes? 
\end{mdframed}
\emph{X Language (Code)}: Give the code answer in another coding language. \\
\textbf{Changed from \ref{code-example:flip-case}}:
Answer the coding question below;
\begin{lstlisting}[language=Python, style=mystyle]
func flipCase(str string) string {
// flipCase takes a string and flips the case of each character: lowercase to uppercase and uppercase to lowercase.

}
\end{lstlisting}
\label{gs:reasoning-style}

\item \emph{Alternative Answer}: Find the alternative solutions to existing solution. \\
\emph{Alternative Answer (Math) }: Give an alternative solution that is different from the standard \textbf{reasoning steps}, but arrives at the same correct final answer.\\
\textbf{Changed from \ref{math-example:john}}: 
\begin{mdframed}[backgroundcolor=pink!20] 
Give an different step-by-step solution to calculate the answer to the following question. Make sure the solution is different from the solution below.
Question: John has 3 boxes. Each box is 5 inches by 6 inches by 4 inches. The walls are 1 inch thick.
What is the total inner volume of all 3 boxes?
Solution:
The walls subtract a (1 + 1) = 2 inches from each dimension. So, each box has a reduced width of (5 - 2) = 3 inches, reduced length of (6 - 2) = 4 inches and reduced height of (4 - 2) = 2 inches. 
So the inner volume of each box is 3 * 4 * 2 = 24 cubic inches. 
The total inner volume of 3 boxes are 3 * 24 = 72 cubic inches.
Alternative Step by Step Solution:
\end{mdframed}
\emph{Alternative Answer (Code)}: Find an alternative solution to existing coding solution.
\textbf{Changed from \ref{code-example:flip-case}}:
\begin{lstlisting}[language=Python, style=mystyle]
Find a different solution other than:
def flip_case(string: str) -> str:

    return string.swapcase()
\end{lstlisting}
\label{gs:alternative-answer}

\item \emph{New Rule:} Integrate a new rule into the original question that requires the solution follow the new rule. This type tests the model's ability to adapt to new ruels and knowledge and use it inside the solution. \\
\emph{Define Rules}:Define a new mathematical rule that will change how the \textbf{toolbox} (commonsense knowledge) is applied during calculation.\\
\textbf{Changed from \ref{math-example:john}:} 
\begin{mdframed}[backgroundcolor=pink!20] 
In a parallel universe, John has 3 boxes. Each box has peculiar dimensions: 5 quarks by 6 quarks by 4 quarks with walls that are 1 quark thick. In this universe, the total inner volume of a box is calculated by using the Illusory Volume operation, represented as IV. The IV operation is defined as: (length * width * height) - (number\_of\_walls * thickness\_of\_each\_wall). What is the total inner volume of all 3 boxes? 
\end{mdframed}
\emph{Simple Name}: The generated code should only have variables names in a certain format. \\
\textbf{Changed from \ref{code-example:flip-case}}:
Answer the coding question below and only use 6 letter word for each variable names inside the solution
\begin{lstlisting}[language=Python, style=mystyle]
def flip_case(string: str) -> str:
    """For a given string, flip lowercase characters to uppercase and uppercase to lowercase.
    >>> flip_case('Hello')
    'hELLO'
    """
\end{lstlisting}
\label{gs:new-rule}
%\end{enumerate}
\end{enumerate}
\end{enumerate}
Overall, these dimensions in the ``Format Change'' and ``Format Constraint'' Domain are designed to challenge the LLMs in ways that reveal their limitations and strengths in maintaining accuracy and functionality under modified or challenging conditions.

\section{Evaluation Details}
\label{app:eval-dets}
We used separate prompt templates for open source and close source models because close source models sometimes give the final answer directly and omit reasoning steps even if prompted with "Let's think step by step".
To ensure the model performs Chain of Thought Reasoning, we use the following prompt template for GPT-4, GPT-3.5, and Gemini to generate the answer:

\begin{mdframed}[backgroundcolor=gray!10]
Solve the question step by step before giving the final answer. Do not directly give the final answer.

\textcolor{blue}{Question}

Reasoning Step:
\end{mdframed}

For Metamath, CodeLlama and Llama2-Chat, we use the following:

\begin{mdframed}[backgroundcolor=gray!10]
Below is an instruction that describes a task. Write a response that appropriately completes the request.

\#\#\# Instruction: \textcolor{blue}{Question}

\#\#\# Response: Let's think step by step.
\end{mdframed}

The temperature of GPT-4 and GPT-3.5 was set to 0.7 (the default setting in OpenAI playground) for \emph{Concept Analysis question} and 0.1 for \emph{Logic Alteration questions and Format Change questions}. Similarly, the temperature for Llama, ChatGPT, and Gemini were set to 0.8 and 0.1 for \emph{Concept Analysis} and \emph{Logic Alteration questions and Format Change questions}, respectively.

\section{Experiment Details}
\label{app:exp-dets}
Prompt to incorporate original answer:
\begin{mdframed}[backgroundcolor=gray!10]
Given the original question and its answer, Solve the question that is a perturbed variant of the original question. Solve the \#perturbed question\# step by step before giving the final answer. Do not directly give the final answer.

\#Original Question\#: \textcolor{blue}{original question}

\#Original Answer\#: \textcolor{blue}{original answer}

\#Perturbed Question\#: \textcolor{blue}{perturbed question}
\end{mdframed}

Self Consistency Prompting:
We randomly picked one question answer pair in the same category from our math dataset \dataset{} and prepend it to the front of the perturbed question as a one shot demonstration. Then we use the below prompt template for Self Consistency prompting. We sample the generation three times and get the final answer by majority voting. In case of tie, we randomly pick an answer.
\begin{mdframed}[backgroundcolor=gray!10]
Given the oneshot demonstration of a question and its final answer, Solve the \#question\# step by step before giving the final answer. Do not directly give the final answer. \\

\#Demonstration Question\#: {\textcolor{blue}{demonstration question}} \\
\#Demonstration Final Answer\#: {\textcolor{blue}{demonstration answer}} \\

\#Question\#: {\textcolor{blue}{question}}
Reasoning Step: [Reasoning Steps]
Final answer: [Final answer]
\end{mdframed}

\textbf{Program of Thought Prompting}:
We use the following prompt template for the program of thought experiments:
\begin{mdframed}[backgroundcolor=gray!10]
Instruction: You are an experienced professional skilled in using python programs to solve math related problems. Solve the question below using python programs, You will only write code blocks. \\
\\
Problem: {\textcolor{blue}{Question}}
\end{mdframed}

\section{Detailed Results}
The detailed results across the perturbation categories for all the models are illustrated in \cref{tab:detailed_results_math,tab:detailed_results_code}.

%\Cref{tab:detailed_results}.
\begin{table*}[ht!]
\centering
\resizebox{\textwidth}{!}{%
\begin{tabular}{l|l|ccccc}
\toprule
\textbf{Dimension} & \textbf{Category} & \textbf{GPT-4} & \textbf{GPT-3.5} & \textbf{Gemini} & \textbf{Metamath} & \textbf{Llama2-Chat} \\
\midrule
Original &                     & 5             & 4                & 4               & 4                 & 3               \\ 
\midrule
\multirow{4}{*}{Granularity Adjustment} & Remove Constraint           & 5             & 5                & 5               & 5                 & 4               \\ 
& Partial Solution              & 5             & 3                & 3               & 3                 & 2               \\ 
& Solution Plan                 & 5             & 4                & 5               & 2                 & 4               \\ 
& Detail Expansion            & 5             & 3                & 5               & 4                 & 0               \\ 
\midrule
\multirow{8}{*}{Reasoning Adjustment} & Add Restriction             & 3             & 1                & 2               & 1                 & 0               \\ 
& Subsequent Question            & 4             & 1                & 3               & 0                 & 0               \\ 
& Concurrent Question           & 4             & 2                & 4               & 1                 & 1               \\ 
& Change Question                & 5             & 2                & 3               & 1                 & 1               \\ 
& Info Recombination            & 4             & 1                & 3               & 0                 & 1               \\ 
& Domain Knowledge              & 4             & 2                & 4               & 2                 & 2               \\ 
& Complex Reality               & 3             & 0                & 1               & 1                 & 0               \\ 
& General Solution              & 5             & 2                & 0               & 0                 & 0               \\
\midrule
\multirow{3}{*}{Computation Adjustment} & Computation Demand          & 4             & 2                & 4               & 1                 & 0               \\ 
& Change Value                 & 1             & 1                & 1               & 1                 & 0               \\ 
& Change Operation              & 5             & 3                & 4               & 1                 & 2               \\
\midrule
\multirow{7}{*}{Symbolic Manipulation} & Symbolic Response           & 4             & 3                & 3               & 0                 & 0               \\ 
& Value Relationship            & 3             & 1                & 2               & 0                 & 0               \\ 
& Variable Group                & 3             & 1                & 2               & 0                 & 0               \\ 
& Backward Reasoning            & 2             & 1                & 1               & 1                 & 0               \\ 
& WhatIf                        & 3             & 1                & 0               & 1                 & 0               \\ 
& Solve Value                   & 5             & 2                & 4               & 1                 & 0               \\ 
& Identify Range                & 1             & 0                & 1               & 1                 & 2               \\
\midrule
\multirow{4}{*}{Question Understanding}& Inherent Premise           & 5             & 2                & 2               & 0                 & 1               \\ 
& Complete Missing              & 5             & 4                & 5               & 2                 & 4               \\ 
& Question Formulation          & 3             & 1                & 2               & 1                 & 1               \\ 
& Add Misinformation            & 4             & 4                & 3               & 3                 & 1               \\
\midrule
\multirow{4}{*}{Solution Evaluation}& Optimize Solution            & 3             & 3                & 2               & 2                 & 4               \\ 
& Step Functionality            & 1             & 0                & 0               & 0                 & 2               \\ 
& Theoretical Basis             & 4             & 4                & 1               & 2                 & 4               \\ 
& Cost Analysis                 & 5             & 2                & 1               & 1                 & 2               \\ 
\midrule
\multirow{5}{*}{Error Debugging}& Seek Clarification           & 1             & 2                & 1               & 0                 & 0               \\ 
& Conditional Analysis          & 3             & 0                & 2               & 0                 & 0               \\ 
& Conflicting Information       & 2             & 0                & 1               & 0                 & 0               \\ 
& Surface Error                 & 4             & 1                & 0               & 1                 & 1               \\ 
& Hidden Error                  & 2             & 0                & 0               & 0                 & 0               \\ 
\midrule
\multirow{4}{*}{Alternative Format} & Setting Rephrase            & 4             & 3                & 2               & 4                 & 1               \\ 
& Change Sequence               & 5             & 2                & 3               & 3                 & 0               \\ 
& Close Format                  & 4             & 2                & 4               & 0                 & 0               \\ 
& Data Restructuring           & 5             & 0                & 2               & 0                 & 0               \\ 
\midrule
\multirow{1}{*}{Pairwise Comparison} & Identical Problem           & 3             & 2                & 1               & 4                 & 3               \\ 
\midrule
\multirow{4}{*}{Answer Constraint} & Reasoning Format            & 4             & 0                & 2               & 0                 & 0               \\ 
& Reasoning Style               & 4             & 0                & 2               & 0                 & 0               \\
& Alternative Answer            & 2             & 0                & 0               & 2                 & 0               \\
& New Rule                      & 3             & 1                & 2               & 2                 & 1               \\
\bottomrule
\end{tabular}
}
\caption{Number of examples correctly predicted by each model on \dataset. There are a total of 5 questions for each category except "Change Value", which only has 2 questions.}
\label{tab:detailed_results_math}
\end{table*}

\begin{table*}[ht!]
\centering
\resizebox{\textwidth}{!}{%
\begin{tabular}{l|l|ccccc}
\toprule
\textbf{Dimension} & \textbf{Category} & \textbf{GPT-4} & \textbf{GPT-3.5} & \textbf{Gemini} & \textbf{CodeLlama} & \textbf{Llama2-Chat} \\
\midrule
Original &  & 4 & 4 & 4 & 3 & 3\\ 
\midrule
\multirow{4}{*}{Granularity Adjustment} & Remove Constraint & 4 & 4 & 4 & 4 & 2\\ 
& Partial Solution & 5 & 3 & 4 & 5 & 2\\ 
& Solution Plan & 5 & 4 & 4 & 3 & 2\\ 
& Detail Expansion & 4 & 3 & 4 & 4 & 3\\ 
\midrule
\multirow{8}{*}{Reasoning Adjustment} & Add Restriction & 0 & 0 & 2 & 2 & 0\\ 
& Subsequent Question & 2 & 2 & 1 & 1 & 3\\ 
& Concurrent Question & 3 & 1 & 0 & 2 & 0\\ 
& Change Question & 2 & 2 & 2 & 2 & 1\\ 
& Info Recombination & 2 & 1 & 1 & 1 & 0\\ 
& Domain Knowledge & 3 & 4 & 3 & 4 & 0\\ 
& Complex Reality & 3 & 2 & 3 & 3 & 0\\ 
& General Solution & 0 & 2 & 1 & 1 & 1\\
\midrule
\multirow{3}{*}{Computation Adjustment} & Computation Demand & 1 & 1 & 2 & 2 & 1\\ 
& Change Value & 2 & 1 & 1 & 2 & 1\\ 
& Change Operation & 4 & 4 & 5 & 2 & 3\\
\midrule
\multirow{7}{*}{Symbolic Manipulation} & Symbolic Response & 4 & 3 & 1 & 2 & 1\\ 
& Value Relationship & 1 & 1 & 1 & 1 & 0\\ 
& Variable Group & 3 & 1 & 1 & 0 & 1\\ 
& Backward Reasoning & 2 & 2 & 3 & 1 & 0\\ 
& WhatIf & 3 & 1 & 0 & 0 & 0\\ 
& Solve Value & 1 & 1 & 0 & 0 & 0\\ 
& Identify Range & 3 & 1 & 2 & 0 & 2\\
\midrule
\multirow{4}{*}{Question Understanding}& Inherent Premise & 2 & 3 & 2 & 1 & 1\\ 
& Complete Missing & 3 & 1 & 3 & 1 & 2\\ 
& Question Formulation & 4 & 4 & 4 & 2 & 3\\ 
& Add Misinformation & 4 & 4 & 4 & 3 & 4\\
\midrule
\multirow{4}{*}{Solution Evaluation}& Optimize Solution & 2 & 2 & 3 & 2 & 2\\ 
& Step Functionality & 5 & 5 & 4 & 2 & 2\\ 
& Theoretical Basis & 5 & 3 & 4 & 0 & 4\\ 
& Cost Analysis & 4 & 5 & 4 & 3 & 2\\ 
\midrule
\multirow{5}{*}{Error Debugging}& Seek Clarification & 2 & 2 & 2 & 3 & 0\\ 
& Conditional Analysis & 1 & 1 & 1 & 0 & 0\\ 
& Conflicting Information & 1 & 0 & 0 & 0 & 0\\ 
& Surface Error & 4 & 4 & 4 & 2 & 1\\ 
& Hidden Error & 3 & 3 & 4 & 2 & 1\\ 
\midrule
\multirow{4}{*}{Alternative Format} & Setting Rephrase & 3 & 2 & 2 & 1 & 3\\ 
& Change Sequence & 4 & 3 & 2 & 3 & 1\\ 
& Close Format & 3 & 1 & 2 & 2 & 0\\ 
& Data Restructuring & 3 & 4 & 3 & 2 & 1\\ 
\midrule
\multirow{1}{*}{Pairwise Comparison} & Identical Problem & 2 & 2 & 2 & 0 & 2\\ 
\midrule
\multirow{4}{*}{Answer Constraint} & Reasoning Format & 2 & 2 & 2 & 2 & 1\\ 
& Reasoning Style & 3 & 2 & 2 & 3 & 1\\
& Alternative Answer & 3 & 3 & 2 & 2 & 0\\
& New Rule & 3 & 2 & 1 & 1 & 2\\
\bottomrule
\end{tabular}
}
\caption{Number of examples correctly predicted by each model on \datasetcode. There are a total of 5 questions for each category.}
\label{tab:detailed_results_code}
\end{table*}

\section{Inclusivity of Skill Set}
\label{sec:skillset}
\paragraph{Dependence between Perturbation Types.}
\begin{table*}[ht!]
\centering
\resizebox{\textwidth}{!}{%
\begin{tabular}{l|l|l|ccccc}
\toprule
Domain & Enhanced Type & Primary Type & GPT-4 & GPT-3.5 & Gemini & Metamath & Llama2-Chat \\
\midrule
\multirow{6}{*}{Logic Alteration} & Backward Reasoning & Solve Value & 20 & 0 & 20 & -20 & 0\\
& Value Relationship & Symbolic Response & 20 & 40 & 40 & 0 & 0 \\
& Variable Group & Symbolic Response & 20 & 40 & 20 & 0 & 0 \\
& Identify Range & Symbolic Response & 60 & 60 & 40 & -20 & -40 \\
& What If & Solve Value & 40 & 20 & 80 & 0 & 0 \\
& Solution Plan & Detail Expansion & 0 & 20 & 0 & -40 & 80 \\
\midrule

\multirow{5}{*}{Concept Analysis} & Seek Clarification & Conditional Analysis & 40 & -40 & 20 & 0 & 0 \\
& Optimize Solution &  Cost Analysis  & 20 & 20 & 20 & 20 & -60 \\
& Conflicting Information & Complete Missing & 20 & 60 & 20 & 40 & 80 \\
& Optimize Solution & Step Functionality & 20 & 0 & 0 & 40 & -40 \\
& Hidden Error & Step Functionality & 0 & 0 & 20 & 0 & 0 \\
\midrule
\multicolumn{3}{c}{Average}  & 25.45 & 16.36 & 30.91 & 0 & 12.73 \\
\bottomrule
\end{tabular}
}
\caption{Performance drop in Enhanced vs. Primary Type questions on \dataset{}. The value equals (accuracy of Primary - accuracy of Enhanced), so positive entries indicate higher performance for Primary Type questions. }
\label{table:type_dependence}
\end{table*}
In our ontology, some specialized perturbation types, which we refer to as \emph{Enhanced types}, require skill in solving some other primary perturbation types which we call \emph{Primary types}. %\hl{[How do you define the skill set? Are the abilities required to solve a parent type a superset of the abilities required to solve its children?]} 
For instance, consider the process of solving perturbed questions generated as outlined in \ref{gs:what-if} The initial step for the model involves identifying the value of an unknown variable from its answer. Subsequently, the model calculates how this value alters the final answer. This initial step demands skills similar to those described in \ref{gs:solve-value} %We detail the reasons for the inclusivity of each skill in parent types over child types in the appendix~\cite{}. 
    Consequently, we anticipate that enhanced perturbation types will be challenging to answer. Following \cref{table:type_dependence}, across all models, \emph{primary types} exhibit higher overall performance as compared to \emph{enhanced types}. Furthermore, it is observed that open-source models do not experience as significant a performance drop as closed-source models when handling enhanced types. This can be attributed to the fact that open-source models already demonstrate near-zero performance in answering primary type questions. Therefore, their inability to answer enhanced questions does not result in a notable decrease in performance. 

\begin{figure*}[ht]
  \centering
  \includegraphics[width=\textwidth]{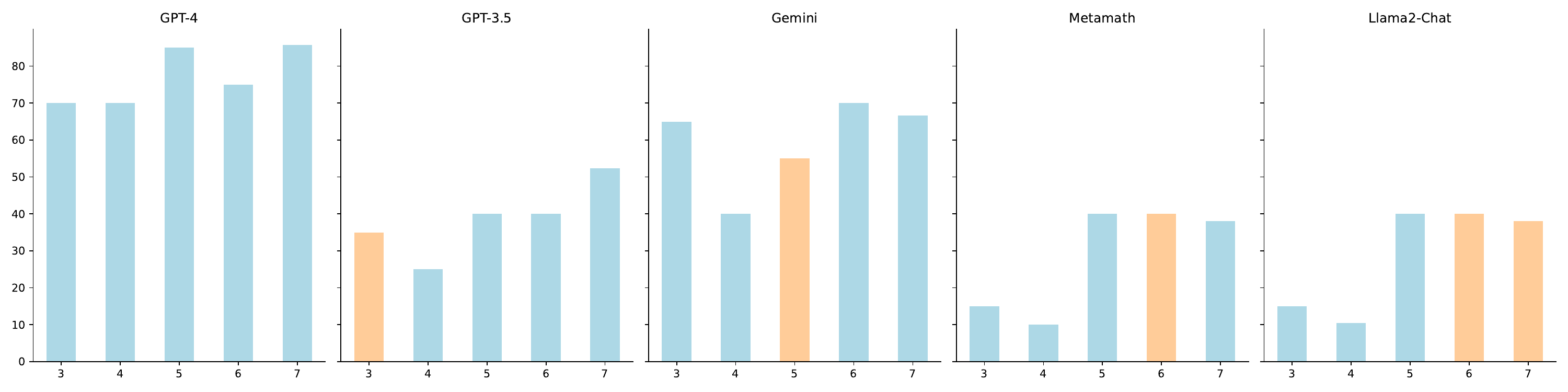}
  \caption{Model performance for each question. The blue color indicates the model predicted correctly for the original question, and orange means the opposite. `3', `4', `5', `7', `8' stands for the number of steps in the gold answer for the perturbed question.}
  \label{fig:cot}
\end{figure*}

\paragraph{Performance across Question Difficulty.}
In our experimentation with various LLMs, we consistently employed the Chain of Thought (CoT) methodology to derive the ultimate answer. This prompts a natural inquiry: \emph{Does the performance of LLMs exhibit any correlation with the number of steps needed to arrive at the final answer?} %\hl{Do we need to state our assumption that if the original question have more CoT steps, the perturbed question will also have more CoT Steps?} 
Surprisingly, in our extensive experiments (as illustrated in \Cref{fig:cot}), we did not discern any definitive correlation or discernible trend. Instead, performance appears to diminish based on the inherent difficulty of the original question in GSM8K. Put differently, if an LLM fails to provide an accurate response to the initial question, its performance similarly falters when confronted with perturbed questions.

\end{document}